%% file: main.tex
\documentclass{article}

\PassOptionsToPackage{numbers, compress}{natbib}
\usepackage[preprint]{neurips_2026}

\usepackage[utf8]{inputenc}
\usepackage[T1]{fontenc}
\usepackage{hyperref}
\usepackage{xurl}
\usepackage{booktabs}
\usepackage{array}
\usepackage{amsmath}
\usepackage{amssymb}
\usepackage{microtype}
\usepackage{xcolor}
\usepackage{graphicx}
\usepackage{caption}
\usepackage{algorithm}
\usepackage{algorithmic}
\usepackage{tikz}
\usepackage{multirow}
\usepackage{wrapfig}
\usepackage{needspace}
\usepackage{placeins}
\usepackage{xspace}
\usepackage{listings}

\lstdefinestyle{prompt}{
  basicstyle=\ttfamily\footnotesize,
  breaklines=true,
  breakatwhitespace=false,
  columns=fullflexible,
  keepspaces=true,
  showstringspaces=false,
  frame=single,
  framerule=0.3pt,
  rulecolor=\color{gray!40},
  backgroundcolor=\color{gray!5},
  xleftmargin=1em,
  xrightmargin=1em,
  breakindent=0pt,
  postbreak=\mbox{\textcolor{gray}{$\hookrightarrow$}\space},
  upquote=true,
  captionpos=b,
}

\lstdefinestyle{pyplan}{
  style=prompt,
  language=Python,
  keywordstyle=\color{blue!70!black},
  commentstyle=\color{green!45!black}\itshape,
  stringstyle=\color{red!60!black},
}

\DeclareMathOperator*{\argmax}{arg\,max}
\newtheorem{definition}{Definition}

\DeclareRobustCommand{\brandname}{\mbox{PrimeScientist}}
\newcommand{\ourmodel}{\textsc{\brandname}\xspace}
\makeatletter
\DeclareRobustCommand{\taskname}[1]{{%
  \sffamily\mdseries\upshape
  \@tempdima=\f@size pt\relax
  \@tempdima=0.92\@tempdima
  \fontsize{\@tempdima}{\baselineskip}\selectfont #1}}
\makeatother
\DeclareRobustCommand{\tasknameinline}[1]{\taskname{\textit{#1}}}

\hypersetup{hidelinks,pdftitle={PrimeScientist: Strategic Allocation of Research Effort in Autonomous Research}}

\title{
\ourmodel:
Strategic Allocation of Research Effort in Autonomous Research
}

\author{%
  Xinle Yu$^{1*}$ \
  Fan Bai$^{2}$ \
  Kaiser Sun$^{2}$ \
  Hengshuo Miao$^{2}$ \ \\
  \textbf{Abhay Anand}$^{1}$ \
  \textbf{Zhongyan Luo}$^{1}$ \
  \textbf{Kun Zhou}$^{1}$ \
  \textbf{Zhen Wang}$^{1*}$ \\
  $^{1}$UC San Diego \
  $^{2}$Johns Hopkins University \\[2pt]
  \texttt{*{zhenwang.work@gmail.com}, xiy033@ucsd.edu}
}

\begin{document}

\makeatletter
\begingroup
\let\originalbottomtitlebar\@bottomtitlebar
\renewcommand{\@bottomtitlebar}{\originalbottomtitlebar\vspace{-12pt}}
\maketitle
\endgroup
\makeatother

\vspace{-22pt}
\begingroup
\setlength{\leftmargini}{18pt}
\begin{abstract}
\vspace{-4pt}
\input{sections/0_abstract}
\end{abstract}
\endgroup

\input{figures/fig_motivation}

\clearpage

\begingroup
\raggedbottom
\input{sections/1_intro}

\input{sections/2_related_work}

\input{sections/3_method}

\input{sections/4_experiments}

\input{sections/6_conclusion}

\Needspace{5\baselineskip}
{\small

\input{main.bbl}
}
\par
\endgroup

\newpage
\appendix
\raggedbottom

\input{sections/7_appendix}

\end{document}

%% file: sections/0_abstract.tex
\setlength{\parfillskip}{0pt plus 0.4\linewidth}
Autonomous research agents aim to automate scientific workflows, from proposing ideas to conducting experiments and analyzing results.
{Yet current AI and research agents can propose more directions than available resources allow them to pursue.}
{Moreover, each attempt could consume substantial resources, requiring agents to reconsider how to invest in subsequent research.}
{Thus, deciding how to invest research effort strategically should be a defining capability of autonomous research agents.}
{Accordingly, we introduce \ourmodel, which jointly determines research direction and resource investment across successive research attempts.}
{Specifically, we formulate this challenge of strategic research effort allocation as a sequential decision problem where remaining resources should explicitly guide the research policy.}
{We first introduce an executable plan tree that preserves competing plans and their outcomes across attempts.}
{Building on this representation, we propose an adaptive MCTS-based allocation policy that balances exploration and exploitation using experimental feedback and remaining resources.}
{Comprehensive evaluations across AI research, systems and code optimization, and machine learning engineering show that strategic allocation improves research quality and sample efficiency together.}
{Across 12 AI research tasks, \ourmodel improves average reward by 10.3\% with 50.6\% fewer research attempts than AutoResearch under the same resource budget.}
{We believe making research effort allocation an explicit optimization target establishes effective resource use as a core research capability for autonomous agents to drive scientific breakthroughs at scale.}\footnote{Code and data are available at \url{https://github.com/Henri-XYu02/PrimeScientist}.}
\par

%% file: figures/fig_motivation.tex
\begingroup
\setlength{\intextsep}{4pt}
\begin{figure}[H]
  \centering
  \includegraphics[width=\textwidth]{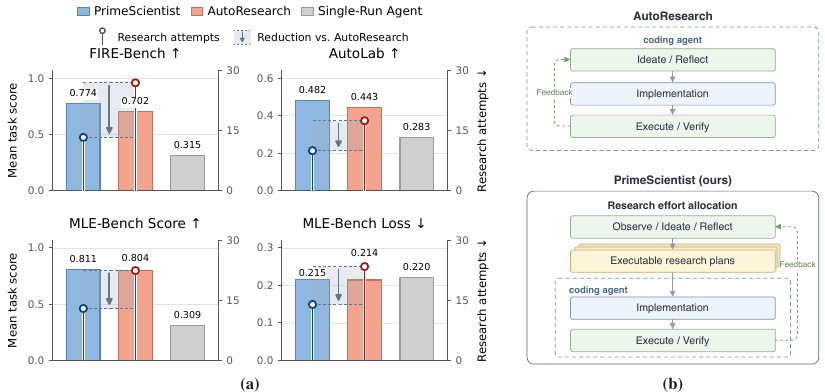}
  \captionsetup{font=small,skip=4pt}
  \caption{\textbf{Strategic research effort allocation {with} \ourmodel.}
  {(a) \ourmodel achieves comparable or better scores with fewer attempts than AutoResearch under matched token budgets. Bars show mean scores; stems and shading show mean attempts and reductions. MLE-Bench separates scores from losses.
  (b) Explicit plans support strategic allocation around the coding agent's execution loop. Experimental feedback and remaining resources guide which research directions receive further effort.}}
  \label{fig:motivation}
\end{figure}
\endgroup

%% file: sections/1_intro.tex
\begingroup
\raggedbottom
\setlength{\parskip}{4.5pt}
\section{Introduction}
\label{sec:intro}
\suppressfloats[t]
\enlargethispage{10pt}

Autonomous research {(using AI agents to} propose hypotheses, modify experimental code, run {experiments}, and decide what to try next) has emerged as a credible regime for both machine-learning engineering and broader scientific work~\cite{autoresearch,ai_scientist,ai_scientist_v2,aide,rd_agent,mlebench}. {In Karpathy's \texttt{autoresearch}~\cite{autoresearch}, a} single agent reads a Markdown plan, edits a Python training file, runs a fixed-time training experiment, and decides whether to keep or discard the change before the next iteration. Around this core loop, a growing line of systems extends the paradigm to ML engineering~\cite{aide,rd_agent,mlebench}, {research ideation and experimental workflows}~\cite{ai_scientist,ai_scientist_v2,researchagent,agent_laboratory}, biomedicine~{\cite{virtuallab,gao2024biomedical,scpilot,cellmaster,t_cell_atlas}}, {chemistry and materials science}~{\cite{coscientist,tritondft}}, and industrial R\&D~\cite{microsoft_discovery}.

{Recent work studies how to use additional computation effectively through inference-time scaling and longer autonomous research loops}~\cite{snell2024scaling,DynScaling,xu2026autolabfrontiermodelssolve}. {Yet agents can generate numerous plausible research ideas and plans, including} alternative hypotheses, dataset choices, model families, ablations, debugging strategies, and interpretations of weak evidence~\cite{researchagent,agent_laboratory,ai_scientist_v2}. Every end-to-end research attempt commits substantial agentic effort (e.g., tokens) to one plan and reduces the opportunities available for subsequent {research attempts}. Progress will depend on how research effort is allocated over time, using both the evidence accumulated so far and the resources that remain~\cite{simon1955behavioral,gershman2015computational,lieder2020resource}. {We therefore study \emph{strategic research effort allocation}{, the problem of deciding} how autonomous agents should distribute effort across research directions as evidence accumulates. The aim is to make effective resource use a deliberate, adaptable part of the research policy. We view this capability as an essential dimension of intelligence in autonomous research agents.}\looseness=-1

\input{tables/tab_research_comparison}

{{However, existing} autoresearch systems rarely treat strategic effort allocation as an explicit capability to optimize.} {AutoResearch} assigns each new attempt to the incumbent trajectory, so the path already taken largely determines where subsequent effort goes~\cite{autoresearch}. Tree- and population-based systems expose multiple alternatives, yet typically navigate them using a fixed search rule or an LLM judgment that does not explicitly represent the remaining research {budget}~\cite{aide,ai_scientist_v2,chi2024selatreesearchenhancedllm,aira_dojo}. R\&D-Agent {selects ideas before implementation and} {adapts planning to the remaining wall-clock time}~\cite{rd_agent}. {Closer to our setting, AlphaLab guides an LLM Strategist using the remaining experiment count, while FML-Bench's AdaptiveSearch switches once from greedy refinement to multi-branch exploration when progress stalls~\cite{alphalab,fmlbench}. {The remaining challenge is to coordinate how research plans evolve and how effort is distributed across them. This calls for a unified policy that uses experimental evidence and remaining resources to guide both planning and execution (Table~\ref{tab:research_comparison}).}}\looseness=-1

{To address this challenge, we introduce \ourmodel to equip autonomous research agents with a policy for directing their research effort. We formulate strategic effort allocation as a sequential decision problem that jointly models plan construction, execution, and their resource costs. To realize this idea, we introduce an executable plan tree that {makes competing plans available for selection before execution.}} A reflector drafts and revises plans from the evidence accumulated in the tree, while a coding-agent executor carries out each selected plan through coding, debugging, experimentation, and result analysis. {Building on this representation, we propose an adaptive MCTS-based allocation policy~\cite{uct,mcts_survey} that uses experimental feedback and remaining resources to balance exploration and exploitation across research directions. Even with the same observed outcomes, different remaining resources can warrant different allocations.}\looseness=-1

We evaluate \ourmodel{} {across AI research, systems and code optimization, and machine learning engineering}~\cite{firebench,xu2026autolabfrontiermodelssolve,mlebench}. \ourmodel achieves {better or comparable} research outcomes with {fewer research attempts than AutoResearch under the same resource budget}. Research sample efficiency measures the task quality obtained relative to the number of complete research attempts. {Each attempt requires implementation, experimentation, and analysis to test a research direction.} Fewer attempts at comparable quality reduce this experimental demand{. The resource budget includes planning and execution; Section~\ref{sec:setup} specifies the units and limits. Our ablation studies show that adapting exploration to the remaining budget improves overall reward over {\emph{UCT}}, {\emph{Greedy}}, {\emph{Random}}, and fixed-exponent alternatives.} {These findings support making strategic effort allocation a core capability of autonomous research agents. Optimizing this capability could enable agents to improve their own research strategies and convert growing computational resources into scientific breakthroughs at scale.}\looseness=-1

\par
\endgroup

%% file: tables/tab_research_comparison.tex
\begin{table}[t]

\centering
\caption{\textbf{Comparison of representative {autonomous research agents}.}
{\emph{Branch exploration} retains alternatives; \emph{plan-level search} selects unexecuted plans between attempts.
\emph{Value backpropagation} updates ancestors from descendant outcomes.
\emph{Strategic effort planning} adjusts exploration and refinement to the remaining budget.
A \emph{shared token budget} covers {all planning and execution tokens.}}}
\label{tab:research_comparison}
\small
\setlength{\tabcolsep}{3.4pt}
\renewcommand{\arraystretch}{1.02}
\begin{tabular}{lccccc}
\toprule
\textbf{Method}
 & \shortstack{\textbf{Branch}\\\textbf{exploration}}
 & {\shortstack{\textbf{Plan-level}\\\textbf{search}}}
 & {\shortstack{\textbf{Value}\\\textbf{backpropagation}}}
 & {\shortstack{\textbf{Strategic effort}\\\textbf{planning}}}
 & {\shortstack{\textbf{Shared token}\\\textbf{budget}}} \\
\midrule
AutoResearch~\cite{autoresearch}          & -- & {--} & {--} & {--} & {--} \\
AIDE~\cite{aide}                         & $\checkmark$ & {--} & {--} & {--} & {--} \\
The AI Scientist-v2~\cite{ai_scientist_v2}& $\checkmark$ & {--} & {--} & {--} & {--} \\
R\&D-Agent~\cite{rd_agent}                & $\checkmark$ & {$\checkmark$} & {--} & {$\checkmark$} & {--} \\
MARS~\cite{mars}                         & $\checkmark$ & {--} & {$\checkmark$} & {--} & {--} \\
AlphaLab~\cite{alphalab}                 & $\checkmark$ & {$\checkmark$} & {--} & {$\checkmark$} & {--} \\
\midrule
\textbf{\ourmodel}                       & $\checkmark$ & {$\checkmark$} & {$\checkmark$} & {$\checkmark$} & {$\checkmark$} \\
\bottomrule
\end{tabular}
\end{table}

%% file: sections/2_related_work.tex
\vspace{-4pt}
\section{Related Work}
\label{sec:related}

\noindent \textbf{{Autonomous Research Agents.}}
Autonomous research agents increasingly automate {workflows covering} ideation, experimental design, implementation, evaluation, and reporting.
\mbox{Systems} such as AI Scientist, ResearchAgent, and Agent Laboratory generate research ideas or artifacts through iterative and multi-stage agentic workflows~{\citep{ai_scientist,researchagent,agent_laboratory,hypoevolve}}.
{Metric-driven systems use task scores to guide progress. For example,} Karpathy's \texttt{autoresearch} repeatedly modifies an experimental program and retains changes that improve the target metric, while AIDE formulates machine-learning engineering as tree search over code solutions~\citep{autoresearch,aide}.
More recent systems make search increasingly explicit.
AI Scientist-v2 uses progressive agentic tree search to develop and evaluate experiments; AIRA compares greedy, MCTS, and evolutionary search policies for machine-learning research; GEAR maintains a population of research states through mutation and crossover; and FML-Bench studies how search topology and exploration dynamics affect research-agent performance~\citep{ai_scientist_v2,aira_dojo,gear,fmlbench}.
Together, these works establish iterative feedback, persistent research state, and structured search as central components of autonomous research.
\ourmodel makes research effort allocation an explicit design choice.
{It studies how empirical evidence and remaining resources should jointly guide the allocation of research attempts.}

\noindent \textbf{{Agent Planning and Search.}}
{Research on language models studies reasoning, planning, and tool use~\citep{rap,llm_reasoners,thinksum,toolkengpt,nabla_reasoner,gpt_turing,modal_mixed_cot,latent_visual_reasoning}. Recent work also makes cost or remaining resources explicit in reasoning and search~\citep{Token-Budget-Aware}.}
MARS introduces cost-constrained MCTS for automated AI research {and backpropagates rewards that combine validation performance with candidate execution time}~\citep{mars}.
AlphaLab {uses a Strategist to revise experiment queues and a playbook to retain experimental lessons. The Strategist uses the remaining experiment count to shift from exploration toward refinement}~\citep{alphalab}.
FML-Bench proposes AdaptiveSearch, which begins with greedy refinement and performs a one-time switch to multi-branch exploration after detecting stagnation, with its branching structure determined by the remaining step budget~\citep{fmlbench}.
{Both approaches adapt trial selection while leaving the resources devoted to planning outside the allocation objective.}
At a finer granularity, BATS gives tool-using agents continuous awareness of remaining tool calls, while BAVT uses the remaining-resource ratio to sharpen a value-based selection distribution from broader exploration toward exploitation within a reasoning tree~\citep{BATS,bavt}.
{These search mechanisms inform our solution, yet our unique contribution is to formulate plan construction and execution as decisions under a shared resource budget. In the executable plan tree, selecting an untested plan initiates a research attempt; selecting an evaluated plan develops alternatives. The same policy therefore directs both research planning and experimentation using experimental evidence and remaining resources.}\looseness=-1

\noindent \textbf{{Resource Allocation in Intelligent Systems.}}
{Resource allocation is a longstanding question in the study of intelligent decision-making. Simon's account of bounded rationality relates effective choice to the information and computational capabilities available to an agent~\citep{simon1955behavioral}. Computational and resource-rational accounts further evaluate reasoning procedures by the quality of their decisions and the resources they require~\citep{gershman2015computational,lieder2020resource}. Rational metareasoning treats the choice of what to compute as a decision in its own right. Recent work applies this principle to language models by learning when intermediate reasoning justifies its computational cost~\citep{rational_metareasoning}.}
In adjacent domains, studies of online innovation tournaments show that solvers strategically vary effort in response to feedback and time remaining, and deadline-aware task-and-motion planning has been formulated as a metareasoned effort-allocation problem over alternative options~\citep{dissanayake2018strategic,sung2024effort}.
{These perspectives make the use of resources part of intelligent decision-making. In autonomous research, agents must construct alternatives and generate the evidence needed to assess them. \ourmodel formalizes the effort devoted to both activities within a sequential decision problem. A research attempt can improve the current result and inform subsequent choices, linking its value to the opportunities for further research.}\looseness=-1

\noindent \textbf{Self-Improving and Self-Evolving Agents.}
A broader line of work enables {language-model systems} to improve their behavior or {revise their underlying procedures}~{\citep{reflexion,self_refine,agrawal2026gepareflectivepromptevolution,promptagent,dynamic_rewarding,self_moe}}.
Reflexion uses verbal feedback and episodic memory to guide subsequent attempts, while GEPA reflects on execution trajectories to evolve prompts and combine complementary lessons~\citep{reflexion,agrawal2026gepareflectivepromptevolution}.
More strongly self-referential systems search over the agent itself.
The Darwin G{\"o}del Machine iteratively modifies its own coding-agent implementation and empirically validates the resulting variants, while Bilevel Autoresearch uses an outer autoresearch loop to generate new {search procedures} for an inner loop~\citep{dgm,bilevel_autoresearch}.
\ourmodel {adapts effort allocation across task-level research plans while retaining the same reflector and executor.}
{Strategic effort allocation could help agents prioritize changes to their own research methods. Optimizing these choices would support recursive self-improvement, with each generation directing its resources toward developing more capable successors.}\looseness=-1

%% file: sections/3_method.tex
\begingroup
\raggedbottom
\setlength{\parskip}{3pt}
\setlength{\abovedisplayskip}{4pt}
\setlength{\belowdisplayskip}{4pt}
\setlength{\abovedisplayshortskip}{3pt}
\setlength{\belowdisplayshortskip}{4pt}
\vspace{-4pt}
\section{{\ourmodel}}
\label{sec:method}

\vspace{-4pt}
\subsection{{The Strategic Research Effort Allocation Problem}}
\label{sec:formulation}

Consider a research task $x$, a plan-construction operator $\mathcal G$, a coding-agent executor $\mathcal A$, a task evaluator $h$, and a resource {budget} $B>0$.
The operator $\mathcal G$ produces executable plans from the task or revises existing plans using accumulated evidence.
The executor carries a selected plan through coding, debugging, experimentation, and analysis, after which $h$ supplies its empirical outcome.
Both operations consume resources; their outputs and realized costs may be unknown before completion.

\vspace{-3pt}
\begin{definition}[Strategic research effort allocation]
\label{def:strategic_effort_allocation}
\normalfont
Let $\mathcal P_t$ denote the available research plans and $\mathcal H_t$ the record of proposals, execution outcomes, and charged costs before decision $t$.
The state comprises these plans, their history, and the remaining resources:
\begin{equation}
    X_t=(x,\mathcal P_t,\mathcal H_t,B-U_t),
    \qquad U_t=\sum_{i<t}c_i,
    \label{eq:allocation_state}
\end{equation}
with $\mathcal P_0=\mathcal H_0=\varnothing$ and $U_0=0$.
A policy $\pi(\cdot\mid X_t)$ chooses an action $a_t$: construct or revise plans through $\mathcal G$, execute a plan $s\in\mathcal P_t$ through $\mathcal A$, or stop.
A construction action can expand $\mathcal P_t$; an execution supplies empirical evidence; either updates $\mathcal H_t$ and incurs cost $c_t$.
Given a terminal research-value functional $\mathcal V$, the objective is
\begin{equation}
    \max_{\pi\in\Pi_B} J_B(\pi),
    \qquad
    J_B(\pi)=\mathbb E_{\pi}\!\left[\mathcal V(\mathcal H_{\tau_\pi})\right],
    \label{eq:allocation_objective}
\end{equation}
where $\tau_\pi$ is the stopping time and $\Pi_B$ contains policies that use only observed history and initiate research actions only while $U_t<B$.
Actions are charged on completion, and no further action starts once the {budget} is reached; the final action may therefore cross the threshold.
The expectation accounts for variability in plan construction and execution.
\end{definition}
\vspace{-3pt}

\input{figures/fig_pipeline}

The central coupling is that {research alternatives must themselves be produced through research effort}.
Constructing a plan changes what can be {tested}, while executing one changes the evidence available for subsequent decisions; both {consume the research budget}.
{The allocation policy jointly chooses which research direction to pursue and whether to construct a plan or execute one.}

\Needspace{3\baselineskip}
\noindent \textbf{{Research sample efficiency.}}
{Research sample efficiency measures the quality of the outcome obtained relative to the number of complete research attempts. We use the best observed task reward as terminal value, $\mathcal V(\mathcal H_\tau)=\max_{i\in\mathcal I_\tau} y_i$, where $\mathcal I_\tau$ indexes completed attempts and $y_i\in[0,1]$; the value is zero if no attempt has been evaluated. The attempt count is $N_\tau=|\mathcal I_\tau|$, counting each executor invocation separately. Comparable value with fewer attempts means that fewer complete cycles of implementation, experimentation, and analysis are needed to obtain the outcome. This matters when testing a research direction requires model training or repeated benchmarking. We assess this {tradeoff between outcomes and effort} under the same inference {budget} $B$, with both planning and execution tokens counted in $U_\tau$. The comparison therefore tests whether resources devoted to choosing research directions translate into more productive research attempts.}

\vspace{-3pt}
\subsection{{Executable Plan Tree for Effort Allocation}}
\label{sec:framework}

{Research effort is directed by choices of hypothesis, experimental design, and implementation strategy. These choices must remain accessible across attempts so that an agent can revisit an earlier direction or develop a distinct alternative. We introduce an executable plan tree $T_t$ to make these choices explicit and assign effort to complete research plans.}

{Each node stores a plan, its recorded outcomes, and branch statistics.} {A reflector constructs these plans; a coding agent executes them.}
{Each plan is a structured Markdown document specifying a research question or proposed change and its implementation strategy (Appendix~\ref{app:skill}).}
An edge $s\to s'$ records a reflector-generated modification of the parent's research strategy.
{Each modification is stored as an executable \texttt{diff.py} together with a description and prior estimate, allowing the child plan to be reconstructed from its parent.}
{Recording the modification alongside parent and child outcomes makes the development of a research strategy traceable and gives the reflector concrete evidence for its next revision.}
Because a plan is stored independently of the executor's conversation, a candidate direction can remain available while another branch is {evaluated}.
For each node $s$, $h(s)$ denotes the reward recorded under \S\ref{sec:setup}; repeated \mbox{executions may yield different outcomes}.
{Figure~\ref{fig:pipeline} (left) shows evaluated and untested plans as distinct nodes.}\looseness=-1

{The tree exposes two research actions: execute an untested plan or expand an evaluated plan through reflection. Node selection therefore determines both the research direction and whether effort develops an alternative or tests one empirically. The tree instantiates the plans and evidence in the allocation state $X_t$, while cumulative token use determines its remaining resources. The executor receives the plan without instructions about the remaining global budget. The reflector can inspect run metadata but receives no additional allocation rule.}\looseness=-1

{The {budget} covers the input and output tokens of both agents, making plan construction compete with execution for the same inference resources.}
\enlargethispage{10pt}
\begin{equation}
    U_t=C_t^{\mathrm{exec}}+C_t^{\mathrm{refl}}.
    \label{eq:shared_budget}
\end{equation}
{Here $B$ measures inference tokens, excluding elapsed time and CPU/GPU hours. Costs convertible to a common unit, such as dollars, fit the same budget formulation; allocation under \mbox{separate resource constraints} remains future work.}

\vspace{-3pt}
\Needspace{4\baselineskip}
\subsection{{Adaptive MCTS-based Effort Allocation}}
\label{sec:instantiation}
\label{sec:reflector}

{Exploring a new direction produces evidence that can improve later research decisions, while refining a promising plan can improve the current result. The balance between these uses of research effort depends on both observed outcomes and the resources available to act on further evidence. We implement this allocation policy within the Monte Carlo tree search (MCTS) framework~\cite{uct,mcts_survey}.}

\noindent \textbf{{Selection.}}
An attempt made early can supply evidence for subsequent allocations; as {resources are consumed}, fewer opportunities remain to act on that evidence.
{The \emph{Select a plan} step in Figure~\ref{fig:pipeline} uses the remaining-budget ratio to adjust how strongly selection favors high-value branches~\cite{bavt}.}

{Selection traverses the tree to an eligible node. Starting from the root, the policy samples a child $v$ of node $u$ with probability proportional to}
{
\begin{equation}
    w_t(v) = \begin{cases}
        Q(v)^{\alpha_t} & v \text{ has been evaluated},\\[2pt]
        \bigl(Q(u)\sqrt{P(v)}\bigr)^{\alpha_t} & v \text{ is unvisited}.
    \end{cases}
    \label{eq:selection_weights}
\end{equation}
}
{Here $Q(v)$ denotes empirical branch value and $P(v)\in[0,1]$ the reflector's prior for an unvisited child. Until a child is evaluated, its weight combines this prior with the parent value $Q(u)$. If $Q(u)=0$, the implementation uses the prior alone, with a positive floor on sampling weights.}

Selection concentration is controlled by
{
\begin{equation}
    \alpha_t=\min\!\left(\frac{1}{r_t},\alpha_{\max}\right),
    \qquad
    r_t=\frac{B-U_t}{B},
    \label{eq:selection_concentration}
\end{equation}
}
with $\alpha_{\max}=10$ in the reported experiments.
{With most resources remaining, selection stays comparatively dispersed. As resources are consumed, larger exponents concentrate selection on higher-value branches.}
Thus, even with the same observed branch values, different remaining budgets induce different allocation distributions.
{Section~\ref{sec:ablations} compares allocation policies with the executor, reflector, tree representation, \mbox{and evaluation procedure held fixed.}}

\noindent \textbf{{Expansion.}}
The reflector first drafts a root plan from the task instruction.
For an evaluated node, it reads the recorded reward, execution evidence, repository state, and relevant tree context, then proposes up to $m$ child plans.
The prompt requests substantive alternatives in hypothesis, implementation strategy, or experimental design.
{Figure~\ref{fig:pipeline} (right) {shows how the observed outcome motivates} a child plan with matched few-shot examples, creating a new direction for subsequent allocation.}
{The child plans enter the tree as separate alternatives for subsequent selection.}
The complete prompt and an example modification are provided in Appendices~\ref{app:reflector_prompt} and~\ref{app:example_plan}.

\noindent \textbf{{Execution and evaluation.}}
{For an untested node $s$, the executor implements the plan, runs experiments, and analyzes the results. The evaluator supplies $h(s)$ under \S\ref{sec:setup}. Task execution provides the evaluation signal directly, without a separate simulation rollout.}

\noindent \textbf{{Backpropagation.}}
{After applying the pruning rule, retained outcomes update values along the selected path.}
The empirical branch value is
{
\begin{equation}
    Q(v)=\frac{1}{|D(v)|}\sum_{s\in D(v)}h(s),
    \label{eq:branch_value}
\end{equation}
}
where $D(v)$ contains retained evaluated descendants of $v$, including $v$ itself once evaluated.
{These updates let subsequent selection use evidence from completed research attempts. Algorithm~\ref{alg:bart} gives the full loop, including pruning and stopping.}

{Together, the plan tree and search policy let \ourmodel revise both its plans and the allocation of effort across them. {Value backpropagation pools evidence across related plans; strategic effort planning adjusts their selection probabilities as resources are consumed.} Effective resource use is thus part of the research policy, linking each attempt to the overall research objective.}\looseness=-1
\par
\endgroup

%% file: figures/fig_pipeline.tex
\begin{figure}[!t]
  \centering
  \includegraphics[width=\linewidth]{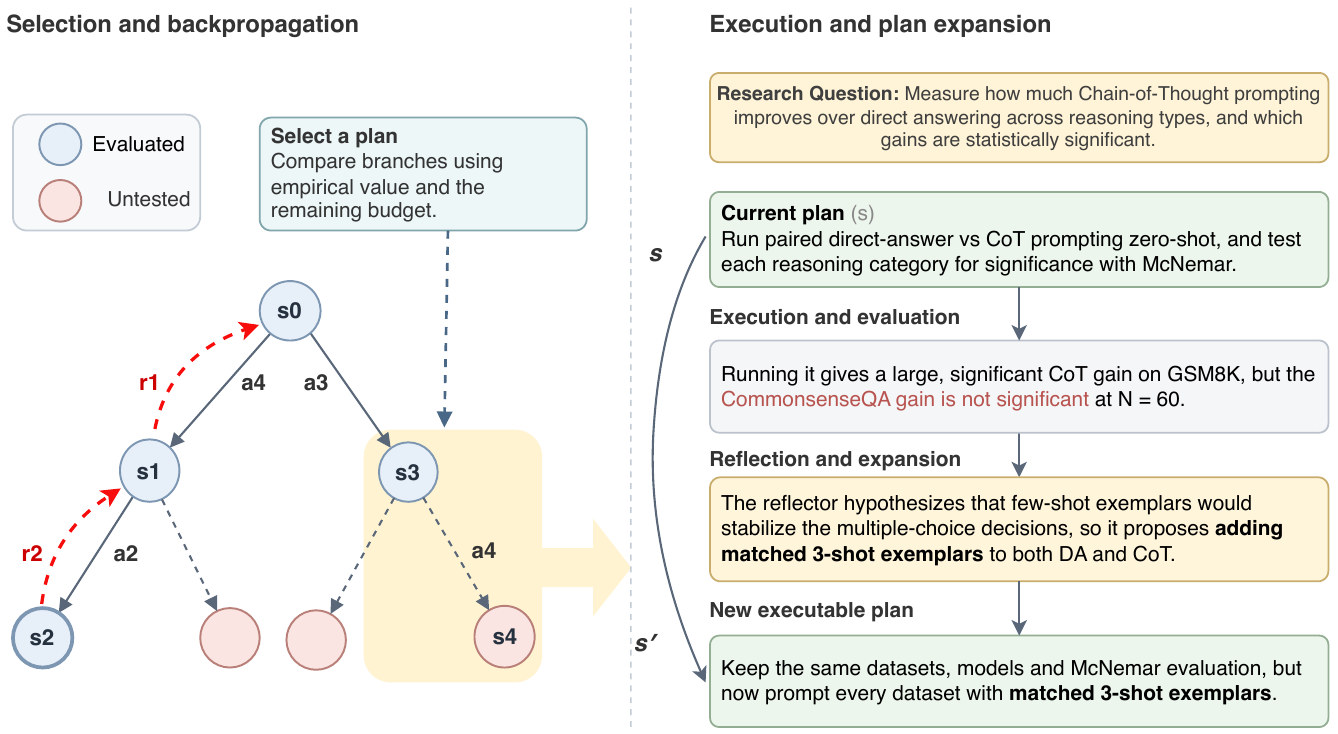}
  \vspace{-2pt}
  \caption{\textbf{The \ourmodel allocation loop.} {The figure shows how \ourmodel organizes research attempts and develops plans from experimental feedback. Plan selection determines both the research direction and whether to develop alternatives or execute a plan. The left panel shows how outcomes update branch values and remaining resources guide subsequent selection. The right panel illustrates how execution feedback motivates a new plan. Planning and execution draw on the same token budget, making both uses of research effort part of the allocation decision.}}
  \label{fig:pipeline}
  \vspace{-11pt}
\end{figure}

%% file: sections/4_experiments.tex
\vspace{-4pt}
\section{Experiments}
\label{sec:experiments}

\vspace{-5pt}
\subsection{Experimental Setup}
\label{sec:setup}

\begingroup
\brokenpenalty=10000
\noindent \textbf{Benchmarks.}
Our evaluation spans three complementary areas of autonomous research: {AI research, systems and code optimization}, and machine learning engineering.
This breadth tests research allocation across experimental design, implementation strategy, and model development.
{\emph{FIRE-Bench}}~\cite{firebench} {evaluates agents on the rediscovery of well-established AI research findings.}
Given a research question, datasets, and experimental requirements, the agent implements experiments and reports conclusions; {the benchmark\textquotesingle s claim-level evaluator, based on RAGChecker}~\cite{ragchecker}, compares these conclusions with reference findings, and we report F1.
{\emph{AutoLab}}~\cite{xu2026autolabfrontiermodelssolve} tests performance optimization under correctness constraints.
We evaluate its eight systems-optimization and puzzle tasks, using throughput-based rewards for the optimization tasks.
{\emph{MLE-Bench}}~\cite{mlebench} evaluates end-to-end machine learning engineering on historical Kaggle competitions.
We report the task-specific scores and metric directions shown in Table~\ref{tab:mlebench}.
{Appendix~\ref{app:research_tasks} describes the tasks across all three benchmarks.}\looseness=-1
\par\endgroup

\noindent \textbf{{Evaluation metrics.}}
{We report the best task score and the number of \emph{research attempts} ({\emph{\#Attempts}}) used under the stated inference {budget}. Task scores follow the benchmark metrics and directions described above. One research attempt is a complete executor invocation covering implementation, experimentation, and analysis; executing the same plan again counts as another attempt. Assessing scores and attempts together measures research sample efficiency{. Achieving comparable quality} with fewer attempts requires fewer complete empirical tests of research directions. Both agents' input and output tokens count toward the {budget}, including the planning used to select those directions. For MLE-Bench, the reported counts cover attempts that produce valid task scores.}
{We compare methods under token budgets, without matching elapsed time or CPU/GPU hours. Other costs could enter the comparison through a common monetary unit; separate resource budgets remain future work.}

\noindent \textbf{Implementation details.}
The executor and reflector use GPT-5 through the Codex CLI, with a shared token {budget} $B=1.5\times10^6$ per task.
{The primary GPT-5 results report one complete search per configuration.}
The main comparisons also impose {limits on research attempts}: 25 on AutoLab and MLE-Bench, and 30 on FIRE-Bench.
Search uses at most $m=3$ child proposals per expansion, exponent cap $\alpha_{\max}=10$, and pruning threshold $\delta=0.05$.
{A child is excluded from subsequent selection if its reward falls more than $\delta$ below its parent's current branch value $Q$, provided $Q>0$;} if all children of a selected node are pruned, reflection can propose further variants.
On FIRE-Bench, each plan node is executed twice and the higher score is recorded, with both executions charged to the budget {and included in the reported attempt counts}.
The reflector normally reads a compressed execution log and consults the full log when necessary; its prompt is provided in Appendix~\ref{app:reflector_prompt}.
A second setting uses GPT-5-mini for both roles and a shared {budget} of $1\times10^6$ tokens.
The AutoLab comparison reports three independent full searches per method and task.
The six-task policy comparison reports three repetitions for \ourmodel's allocation policy and two for each alternative, including a fixed exponent $\alpha=3$.

\input{tables/tab_autolab}

\begingroup
\brokenpenalty=10000
\noindent \textbf{Baselines.}
We compare three baseline families using the same coding-agent backbone within each setting.
\emph{(B1) Single-run Agent} measures execution capability with one executor invocation on the initial workspace, without reflection.
\emph{(B2) AutoResearch}~\cite{autoresearch} evaluates linear iterative refinement through an edit-run-keep-or-revert loop{. Subsequent iterations develop the current trajectory using its history to guide the next invocation.}
It has no reflector-proposed branching or tree pruning.
The comparison between \ourmodel and AutoResearch evaluates the complete allocation framework, including its reflection overhead.
\emph{(B3) Alternative allocation policies} retain the tree-based framework and replace its allocation rule with {\textit{Random}}, {\textit{Greedy}}, {\textit{UCT}}, or fixed-exponent sampling (\S\ref{sec:ablations}), isolating the effect of research allocation.
For AutoLab, the column labeled \emph{AutoLab} additionally reports the benchmark's published baseline as an external reference.
\par
\endgroup

\input{tables/tab_firebench}
\input{tables/tab_mlebench}

\subsection{Main Results}
\label{sec:main_results}

\noindent \textbf{{Higher AI research quality with fewer attempts.}}
{On FIRE-Bench (Table~\ref{tab:firebench}), \ourmodel obtains an average reward of $0.7738$, compared with $0.7018$ for AutoResearch, while reducing research attempts from $27.0$ to $13.33$. Across the twelve tasks, this corresponds to $10.3\%$ higher average reward with $50.6\%$ fewer attempts. Rewards improve on six tasks and match on two, with fewer attempts on all twelve. The gains extend beyond {\tasknameinline{MCQ Selection Bias}}, the largest individual improvement{. Excluding that task} leaves mean rewards of $0.7533$ versus $0.7266$. \mbox{On {\tasknameinline{LLM Value Consistency}}}, for instance, reward rises from $0.727$ to $0.889$ with $14$ {versus} $30$ attempts. These results support strategic allocation as a means to improve research quality and sample efficiency together.}\looseness=-1

\noindent \textbf{{Comparable optimization quality with fewer research attempts.}}
{On AutoLab, \ourmodel achieves an average reward of $0.4824$, compared with $0.4428$ for AutoResearch, using $10.0$ versus $17.5$ research attempts {(Table~\ref{tab:autolab})}. It obtains a nonzero reward on the accuracy-gated {\tasknameinline{Smallest Game Player}} task. On the remaining seven tasks, average rewards are comparable ($0.5052$ versus $0.5061$), with fewer attempts on every task. The benefit therefore includes preserving optimization quality while reducing experimentation, as well as finding stronger solutions{. On} {\tasknameinline{Concurrent KV WAL}}, reward improves from $0.5946$ to $0.6335$ with $10$ versus $18$ attempts. These results extend strategic allocation to implementation choices tested under correctness constraints.}

\noindent \textbf{{Sample efficiency extends to machine learning engineering.}}
{On MLE-Bench (Table~\ref{tab:mlebench}), task scores are close and each method leads on two of the four competitions. \ourmodel achieves higher scores on {\tasknameinline{APTOS 2019 Blindness}} and {\tasknameinline{Plant Pathology 2020}}, while AutoResearch attains slightly lower losses on {\tasknameinline{NOMAD 2018}} and {\tasknameinline{Spooky Author ID}}. Across all four, \ourmodel uses fewer research attempts with valid task scores{, using $11$ to $15$ per task, compared with $21$ to $25$.} These results extend the sample-efficiency benefit to research attempts \mbox{involving data processing, training, and validation.}}\looseness=-1

\noindent \textbf{{Strategic allocation makes research attempts more productive.}}
{Across the three benchmarks, \ourmodel uses fewer research attempts on $23$ of $24$ tasks and achieves higher or comparable benchmark-level performance. Planning and execution both count toward the token budget, so the comparison includes the inference spent on choosing research directions. Taken together, the results show that allocating part of the research effort to planning can preserve or improve outcomes while reducing the number of attempts needed to test research directions. Section~\ref{sec:reproducibility} examines whether this benefit persists across independent searches.}
{Appendices~\ref{app:resource_accounting}, \ref{app:matched_resources}, and~\ref{app:runtime} provide the token breakdown, reward comparisons at matched resources, and elapsed search times.}

\subsection{{Effect of Budget Adaptivity}}
\label{sec:ablations}

{Effective research allocation requires deciding how much effort to devote to testing alternatives and developing promising plans. This balance can change as experiments produce new evidence and resources are consumed. We examine how the selection rule affects research quality, then test whether adapting exploration to the remaining budget improves the outcome.}

{We isolate the effect of plan selection by varying only the allocation policy, with the executor, reflector, plan tree, pruning rule, and budget accounting held fixed.}
In the GPT-5 experiments, {\textit{Random}} samples children uniformly, {\textit{Greedy}} selects the highest-valued child, and {\textit{UCT}}{~\cite{uct}} uses the conventional exploration bonus with $c=\sqrt{2}$.
The eight-task subset contains four tasks from FIRE-Bench and four from AutoLab, listed in Table~\ref{tab:ablation}.
\looseness=-1

\input{tables/tab_ablation}
\input{tables/tab_policy_sensitivity}
\input{tables/tab_autolab_stability}
\input{figures/fig_sample_efficiency}

\noindent \textbf{{Adaptive plan selection improves overall reward.}}
Table~\ref{tab:ablation} reports an average reward of $0.598$ for \ourmodel's policy, compared with $0.551$ for {\textit{UCT}}, $0.563$ for {\textit{Greedy}}, and $0.546$ for {\textit{Random}}.
{Only the allocation policy changes in this comparison, linking the higher overall reward to how research effort is distributed across plans.}
{\ourmodel exceeds {\textit{UCT}} on seven tasks and matches it on the eighth. The gain therefore extends across the evaluated subset, even though both policies use the same plan representation and {reflector}.}
{{\textit{Random}} leads on {\tasknameinline{SECA Hallucination}} and {\tasknameinline{Gaussian Blur}}, showing that the best policy can differ across tasks.}\looseness=-1

\noindent \textbf{{Adapting exploration to the remaining budget improves reward.}}
{The fixed-exponent comparison directly tests the benefit of budget adaptivity.} {Using GPT-5-mini, we compare} \ourmodel's allocation rule with three constant exponents: $\alpha=0$ (uniform sampling), $\alpha=3$, and $\alpha=10$ (strongly concentrated sampling, labeled {\textit{Greedy}}).
{The research framework and sampling rule remain fixed; {the exponent either adapts to the remaining budget or stays constant}.}
{This finite-exponent policy differs from deterministic {\textit{Greedy}} selection in Table~\ref{tab:ablation}.}\looseness=-1

{Adapting the selection exponent achieves the strongest overall reward without choosing a separate constant for each task.}
Table~\ref{tab:policy_sensitivity} reports the highest overall mean for \ourmodel's policy ($0.537$), followed by the fixed $\alpha=10$ policy ($0.507$), uniform sampling ($0.502$), and fixed $\alpha=3$ ($0.452$).
{Our policy leads on three of six tasks, exceeds fixed $\alpha=3$ on all six, and never ranks last.} {Uniform sampling leads on {\tasknameinline{QuestBench}}, while $\alpha=10$ leads on {\tasknameinline{LLM Value Consistency}}.}\looseness=-1

\Needspace{6\baselineskip}
\vspace{-5pt}
\subsection{{Consistency across Runs and Tasks}}
\label{sec:reproducibility}

{Sample-efficiency gains should persist across independent searches and different task requirements. We examine this consistency using three searches per method and task with GPT-5-mini on AutoLab (Table~\ref{tab:autolab_stability}). Alongside reward and research attempts, we inspect accuracy on the task whose reward threshold can obscure improvements in the underlying solution.}

\noindent \textbf{{Comparable reward with fewer research attempts.}}
Across the six tasks included in the average, Table~\ref{tab:autolab_stability} reports mean rewards of $0.4363$ for \ourmodel and $0.4359$ for AutoResearch.
{{The mean number of research attempts falls} from ${28.4}$ to $21.6$, a reduction of approximately $24\%$.}
{The average attempt count is lower on every task across three independent searches per method and task, supporting a consistent sample-efficiency benefit across this evaluation.}
{On {\tasknameinline{Concurrent KV WAL}}, \ourmodel improves reward ($0.5830$ versus $0.5608$) while reducing attempts ($13.0$ versus $25.7$). On {\tasknameinline{Hash Join}}, {\tasknameinline{Flash Attention}}, and {\tasknameinline{FFT (Rust)}}, it achieves comparable reward with fewer attempts. These results show that strategic allocation can preserve task performance while reducing the number of research attempts across distinct optimization problems. Appendix~\ref{app:budget_sensitivity} further examines how these benefits vary with the total token {budget}.}

\noindent \textbf{{Higher accuracy with fewer research attempts.}}
{On {\tasknameinline{Smallest Game Player}}, \ourmodel achieves a higher mean maximum accuracy ($0.926$ versus $0.918$) with fewer attempts ($12.3$ versus $22.0$). Table~\ref{tab:autolab_stability} reports accuracy separately to expose progress below the task's $0.95$ reward threshold.}
{Across the three searches, \ourmodel's maximum accuracy ranges from $0.916$ to $0.940$, compared with $0.872$ to $0.944$ for AutoResearch. The higher minimum accuracy complements the higher average, showing more consistent progress toward the correctness threshold in this setting. Reporting accuracy makes this progress visible even when the gated reward remains zero.}

\par\begingroup
\Needspace{8\baselineskip}
\vspace{-4pt}
\subsection{{Analysis of Research Quality versus Research Effort}}
\label{sec:allocation_trajectories}

{We examine how many research attempts are needed to reach a reference score and how research quality evolves as input tokens accumulate.}

\input{figures/fig_llm_value_consistency}
\rightskip=0pt plus 1em
\indent\noindent \textbf{{Fewer attempts to reach reference scores.}}
{Figure~\ref{fig:sample_efficiency} measures progress toward each method's reference score, set to $90\%$ of its competitor's best result on the task.}
\ourmodel reaches its corresponding threshold in fewer {attempts} on {five of the six tasks}.
{The comparison shows how much experimentation each method requires to approach the other's final research quality.}

\noindent \textbf{{Gains after AutoResearch plateaus.}}
{\mbox{Figure~\ref{fig:llm_value_consistency}} traces a single search on {\tasknameinline{LLM Value Consistency}}. {This task studies whether LLM responses express consistent human values across different contexts and question framings~\citep{rozen2025values}.}}
\ourmodel's running-best score continues to improve after AutoResearch plateaus, reaching approximately $0.89$ versus $0.73$.
{The search budget includes \mbox{both input and output tokens.}}

\par\WFclear
\rightskip=0pt
\endgroup

\Needspace{6\baselineskip}
\subsection{{Case Study of Strategic Effort Allocation}}
\label{sec:casestudy}

{To understand how strategic allocation shapes the experiments an agent performs, we examine {\tasknameinline{Learning Order Agreement}}. {Independently trained networks often learn to classify the same images earlier than others~\citep{hacohen2020agree}. The task asks what explains this shared learning order. Figure~\ref{fig:case_study_loa} shows how \ourmodel tests competing explanations and chooses which plans to develop further.}}

\input{figures/fig_case_study_loa}

\Needspace{3\baselineskip}
\noindent \textbf{{Executable plans test alternative explanations.}}
{The reflector begins with repeated CNN training on CIFAR-10, including pixel-permutation and random-label comparisons. The coding agent \mbox{executes} this plan and reports findings, scoring F1 $0.689$. The reflector then proposes label-noise and architecture experiments to test alternative explanations for shared learning order.}

\noindent \textbf{{Selective expansion yields stronger research findings.}}
{Both branches develop further experiments. The label-noise branch examines how different patterns of incorrect labels affect learning order, while a lower-scoring Mixup extension is pruned. The architecture branch alters image structure through patch shuffling and Fourier phase scrambling, reaching F1 $0.857$. A separate optimizer and batch-size comparison remains unexecuted. The search thus explores different explanations while selectively committing effort to their experimental tests.}

{In this case, \ourmodel matches AutoResearch's research quality with fewer attempts. Both reach F1 $0.857$, using $12$ and $30$ attempts, respectively (Table~\ref{tab:firebench}). The tree makes this allocation visible through the alternatives developed, pruned, or left unexecuted.}

%% file: tables/tab_autolab.tex
\begin{table}[t]
\centering
\caption{{\textbf{Systems and code optimization on AutoLab.} \ourmodel achieves higher overall reward with fewer research attempts than AutoResearch. Each configuration uses one GPT-5 search with a 1.5M-token budget. The AutoLab columns give the benchmark's published baseline.}}
\label{tab:autolab}
\setlength{\tabcolsep}{4pt}
\resizebox{\textwidth}{!}{
\begin{tabular}{lcccccccc}
\toprule
 & \multicolumn{2}{c}{\textbf{\ourmodel}} & \multicolumn{2}{c}{\textbf{AutoResearch}} & \multicolumn{2}{c}{\textbf{Single-run Agent}} & \multicolumn{2}{c}{\textbf{AutoLab}} \\
\cmidrule(lr){2-3}\cmidrule(lr){4-5}\cmidrule(lr){6-7}\cmidrule(lr){8-9}
\textbf{Task} & \textbf{Max reward} & \textbf{{\#Attempts}} & \textbf{Max reward} & \textbf{{\#Attempts}} & \textbf{Max reward} & \textbf{{\#Attempts}} & \textbf{Max reward} & \textbf{{\#Attempts}} \\
\midrule
{\taskname{Gaussian Blur}}        & 0.5122          &  10 & 0.5165 & 20 & 0.0000    & 1 & 0.1376 & 1\\
{\taskname{Hash Join}}            & 0.6306          & 12 & 0.6376 & 16 &   0.6019   & 1 & 0.5484 & 1\\
{\taskname{Concurrent KV WAL}}    & 0.6335         &  10 & 0.5946 & 18 & 0.4844     & 1 & 0.4824 & 1 \\
{\taskname{Flash Attention}}      & 0.3572          & 11 & 0.3781 & 16 & 0.2029     & 1 & 0.2240 & 1\\
{\taskname{FFT (Rust)}}           & 0.5192 & 14 & 0.5317 &  25 & 0.5002     & 1 & 0.4444 & 1\\
{\taskname{VLIW Scheduler}}       & 0.5350 &  9 & 0.5350   & 25 & 0.4620     & 1 & 0.5965 & 1\\
{\taskname{Smallest Game Player}} & 0.3224 &  8 & 0.0000 &  7 & 0.0000     & 1 & 0.0000 & 1\\
{\taskname{Sha256 Throughput}} & 0.3490 & 6 & 0.3491 & 13 & 0.0106 & 1 & 0.0000 & 1\\
\midrule
\textbf{Average}     & 0.4824 & 10.0 & 0.4428 & 17.5 & 0.2828 & 1 & 0.3042 & 1\\\bottomrule
\end{tabular}}
\end{table}

%% file: tables/tab_firebench.tex
\begin{table}[t]
\centering
\caption{{\textbf{AI research on FIRE-Bench.} \ourmodel achieves higher research quality with about half as many attempts as AutoResearch. Results report best F1 from one GPT-5 search per configuration (1.5M tokens). \emph{\#Attempts} includes both executions per \ourmodel plan.}}
\label{tab:firebench}
\setlength{\tabcolsep}{4pt}
\resizebox{\textwidth}{!}{
\begin{tabular}{lcccccc}
\toprule
 & \multicolumn{2}{c}{\textbf{\ourmodel}} & \multicolumn{2}{c}{\textbf{AutoResearch}} & \multicolumn{2}{c}{\textbf{Single-run Agent}} \\
\cmidrule(lr){2-3}\cmidrule(lr){4-5}\cmidrule(lr){6-7}
\textbf{Task} & \textbf{Max reward} & \textbf{{\#Attempts}} & \textbf{Max reward} & \textbf{{\#Attempts}} & \textbf{Max reward} & \textbf{{\#Attempts}} \\
\midrule
{\taskname{Activation Control}}       & 0.3810 & 10 & 0.4400 & 24 & 0.2220 & 1 \\
{\taskname{SECA Hallucination}}       & 0.6980 & 16 & 0.6320 & 30 & 0.6320 & 1 \\
{\taskname{QuestBench}}               & 0.6820 & 16 & 0.7500 & 25 & 0.4000 & 1 \\
{\taskname{Learning Order Agreement}} & 0.8570 & 12 & 0.8570 & 30 & 0.8000 & 1 \\
{\taskname{LLM Value Consistency}} & 0.8890 & 14 & 0.7270 & 30 & 0.3330 & 1 \\
{\taskname{To CoT or not to CoT}}      & 0.8570 & 16 & 0.7740 & 27 & 0.6000 & 1 \\
{\taskname{Grokking or Not}}           & 0.4900 & 14 & 0.5000 & 23 & 0.0000 & 1 \\
{\taskname{Max Suppression}}           & 0.7740 & 16 & 0.7060 & 25 & 0.0000 & 1 \\
{\taskname{Neural Collapse Losses}}    & 1.0000 & 4 & 1.0000 & 25 & 0.0000 & 1 \\
{\taskname{CoT Faithfulness Gaps}} & 0.8890 & 16 & 0.7500 & 29 & 0.0000 & 1 \\
{\taskname{Lost in The Middle}} & 0.7690 & 14 & 0.8570 & 30 & 0.6450 & 1\\
{\taskname{MCQ Selection Bias}} & 1.0000 & 12 & 0.4290 & 26 & 0.1430 & 1\\

\midrule
\textbf{Average}         & 0.7738 & 13.33 & 0.7018 & 27.0 & 0.3146 & 1 \\
\bottomrule
\end{tabular}}
\end{table}

%% file: tables/tab_mlebench.tex
\begin{table}[t]
\centering
\caption{{\textbf{Machine learning engineering on MLE-Bench.} \ourmodel achieves comparable scores with fewer research attempts. Best scores come from one GPT-5 search per configuration (1.5M tokens). \emph{\#Attempts} counts research attempts that return a valid task score.}}
\label{tab:mlebench}
\setlength{\tabcolsep}{4pt}
\resizebox{\textwidth}{!}{
\begin{tabular}{lcccccc}
\toprule
 & \multicolumn{2}{c}{\textbf{\ourmodel}} & \multicolumn{2}{c}{\textbf{AutoResearch}} & \multicolumn{2}{c}{\textbf{Single-run Agent}} \\
\cmidrule(lr){2-3}\cmidrule(lr){4-5}\cmidrule(lr){6-7}
\textbf{Task} & \textbf{Best score} & \textbf{{\#Attempts}} & \textbf{Best score} & \textbf{{\#Attempts}} & \textbf{Best score} & \textbf{{\#Attempts}}  \\
\midrule
{\taskname{APTOS 2019 Blindness}} (↑) & 0.8017 & 11 & 0.7922 & 21 & 0.1004 & 1\\
{\taskname{NOMAD 2018}} (↓) & 0.0597 & 15 & 0.0591 & 25 & 0.0637 & 1\\
{\taskname{Plant Pathology 2020}} (↑) & 0.8208 & 15 & 0.8149 & 24 & 0.5185 & 1\\
{\taskname{Spooky Author ID}} (↓) & 0.3704 & 13 & 0.3698 & 22 & 0.3767 & 1\\
\bottomrule
\end{tabular}}
\end{table}

%% file: tables/tab_ablation.tex
\begin{table}[t]
\centering
\caption{{\textbf{Adaptive allocation improves research outcomes.} The \ourmodel policy yields the highest overall reward when only plan selection changes. All variants share the executor, reflector, plan tree, pruning rule, and 1.5M-token budget. Each configuration uses one GPT-5 search.}}
\label{tab:ablation}
\setlength{\tabcolsep}{6pt}
\begin{tabular}{llcccc}
\toprule
\textbf{Benchmark} & \textbf{Task} & \textbf{\ourmodel} & \textbf{UCT} & \textbf{Greedy} & \textbf{Random} \\
\midrule
\multirow{4}{*}{FIRE-Bench}
 & {\taskname{Activation Control}}       & 0.381 & 0.353          & 0.279          & 0.381 \\
 & {\taskname{LLM Value Consistency}}   & 0.889 & 0.889 & 0.889 & 0.727          \\
 & {\taskname{SECA Hallucination}}       & 0.698          & 0.585          & 0.686          & 0.750 \\
 & {\taskname{QuestBench}}                & 0.682 & 0.649          & 0.600          & 0.545          \\
\midrule
\multirow{4}{*}{AutoLab}
 & {\taskname{Concurrent KV WAL}}       & 0.633          & 0.618          & 0.635 & 0.517          \\
 & {\taskname{Flash Attention}}          & 0.357 & 0.304          & 0.301          & 0.341          \\
 & {\taskname{Hash Join}} & 0.631 & 0.619 & 0.625 & 0.591 \\
 & {\taskname{Gaussian Blur}} & 0.512 & 0.394 & 0.488 & 0.516\\
 
\midrule
\multicolumn{2}{l}{\textbf{Average} (8 tasks)} & 0.598 & 0.551 & 0.563 & 0.546 \\
\bottomrule
\end{tabular}
\end{table}

%% file: tables/tab_policy_sensitivity.tex
\begin{table}[t]
\centering
\caption{{\textbf{Adapting exploration improves reward over fixed policies.} \ourmodel adapts the sampling exponent to the remaining budget; the alternatives hold it constant. Both agents use GPT-5-mini with a shared 1M-token budget. Entries give mean maximum reward (population standard deviation) over three independent searches for \ourmodel and two per fixed policy.}}
\label{tab:policy_sensitivity}
\setlength{\tabcolsep}{6pt}
\resizebox{\textwidth}{!}{
\begin{tabular}{llcccc}
\toprule
\textbf{Benchmark} & \textbf{Task} & \textbf{\ourmodel} & \textbf{Random} ($\alpha{=}0$) & \textbf{Fixed} ($\alpha{=}3$) & \textbf{Greedy} ($\alpha{=}10$) \\
 & & \scriptsize{(n=3)} & \scriptsize{(n=2)} & \scriptsize{(n=2)} & \scriptsize{(n=2)} \\
\midrule
\multirow{3}{*}{FIRE-Bench}
 & {\taskname{Activation Control}}     & 0.314 (0.010) & 0.245 (0.034)          & 0.203 (0.016) & 0.279 (0.045) \\
 & {\taskname{QuestBench}}              & 0.555 (0.085)          & 0.664 (0.064) & 0.467 (0.005) & 0.431 (0.031) \\
 & {\taskname{LLM Value Consistency}} & 0.850 (0.108)          & 0.856 (0.067)          & 0.656 (0.071) & 0.962 (0.038) \\
\midrule
\multirow{3}{*}{AutoLab}
 & {\taskname{Concurrent KV WAL}}     & 0.583 (0.020) & 0.369 (0.084)          & 0.499 (0.018) & 0.451 (0.196) \\
 & {\taskname{Hash Join}}              & 0.626 (0.003) & 0.582 (0.029)          & 0.617 (0.002) & 0.624 (0.008) \\
 & {\taskname{Flash Attention}}        & 0.294 (0.037)          & 0.294 (0.030)          & 0.270 (0.030) & 0.297 (0.088) \\
\midrule
\multicolumn{2}{l}{\textbf{Average} (6 tasks)} & 0.537 & 0.502 & 0.452 & 0.507 \\
\multicolumn{2}{l}{\textbf{best (highest mean) per task}} & 3 / 6 & 1 / 6 & 0 / 6 & 2 / 6 \\
\bottomrule
\end{tabular}}

\end{table}

%% file: tables/tab_autolab_stability.tex
\begin{table}[t]
\centering
\caption{\textbf{Comparable reward with fewer research attempts on AutoLab.} {Three independent GPT-5-mini searches per configuration (1M tokens each) test consistency. Rewards are mean $\pm$ population standard deviation; \emph{\#Attempts} is the mean count. The separate \tasknameinline{Smallest Game Player} row reports maximum accuracy as mean (range), showing progress below its $0.95$ reward threshold.}}
\label{tab:autolab_stability}
\setlength{\tabcolsep}{4pt}
\resizebox{\textwidth}{!}{
\begin{tabular}{lcccc}
\toprule
 & \multicolumn{2}{c}{\textbf{\ourmodel}} & \multicolumn{2}{c}{\textbf{AutoResearch}} \\
\cmidrule(lr){2-3}\cmidrule(lr){4-5}
\textbf{Task} & \textbf{Max reward (mean $\pm$ std)} & \textbf{{\#Attempts}} & \textbf{Max reward (mean $\pm$ std)} & \textbf{{\#Attempts}} \\
\midrule
{\taskname{Gaussian Blur}}        & 0.4740 $\pm$ 0.0487\ & 25.7 & 0.4859 $\pm$ 0.0145\ & 29.3 \\
{\taskname{Hash Join}}             & 0.6259 $\pm$ 0.0029\  & 27.0 & 0.6291 $\pm$ 0.0050\  & 36.0 \\
{\taskname{Concurrent KV WAL}}    & 0.5830 $\pm$ 0.0203\  & 13.0 & 0.5608 $\pm$ 0.0377\  & 25.7 \\
{\taskname{Flash Attention}}      & 0.2939 $\pm$ 0.0372\  & 17.3 & 0.2960 $\pm$ 0.0113\  & 19.3 \\
{\taskname{FFT (Rust)}}            & 0.5002 $\pm$ 0.0037\  & 30.7 & 0.5133 $\pm$ 0.0151\  & 41.0 \\
{\taskname{Sha256 Throughput}}    & 0.1410 $\pm$ 0.1523\  & 16.0 & 0.1304 $\pm$ 0.1540\ & 19.3 \\
\midrule
\textbf{Average} (6 tasks) & 0.4363 & 21.6 & 0.4359 & {28.4} \\
\midrule
\multicolumn{5}{l}{\emph{Accuracy-gated task (excluded from the average); maximum accuracy: mean (range).}} \\
{\taskname{Smallest Game Player}} & 0.926\ \ (0.916--0.940) & 12.3 & 0.918\ \ (0.872--0.944) & 22.0 \\
\bottomrule
\end{tabular}}
\vspace{12pt}
\end{table}

%% file: figures/fig_sample_efficiency.tex
\begin{figure}[!t]
  \centering
    \begin{tikzpicture}
      \node[anchor=south west,inner sep=0] (plot) at (0,0)
        {\includegraphics[trim=7bp 9bp 7bp 6bp,clip,width=\linewidth]{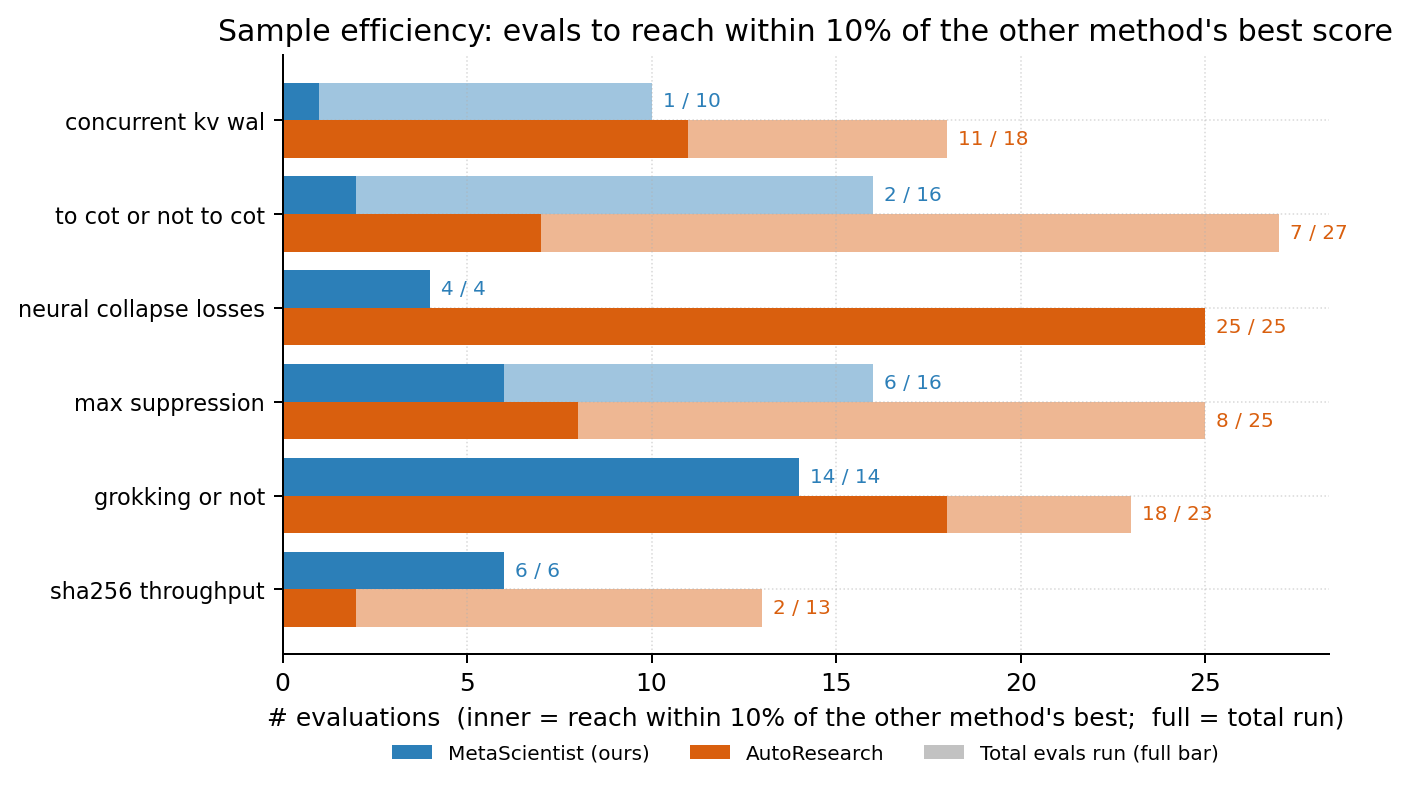}};
      \begin{scope}[x={(plot.south east)},y={(plot.north west)}]
        \fill[white] (0.310,0.000) rectangle (0.475,0.042);
        \node[anchor=base west,inner sep=0,text=black,
          font=\fontfamily{phv}\fontsize{6.4}{7.4}\selectfont]
          at (0.315,0.0105) {\brandname{} (ours)};
        \fill[white] (0.140,0.953) rectangle (1.000,1.000);
        \node[anchor=center,inner sep=0,text=black,
          font=\fontfamily{cmss}\fontseries{m}\fontshape{n}\fontsize{9}{10}\selectfont]
          at (0.570,0.975) {Fewer attempts to approach the other method's best score};
        \fill[white] (0.170,0.043) rectangle (1.000,0.092);
        \node[anchor=center,inner sep=0,text=black,
          font=\fontfamily{phv}\fontsize{8}{9}\selectfont]
          at (0.575,0.067) {Research attempts};
        \fill[white] (0.690,0.000) rectangle (0.905,0.042);
        \node[anchor=base west,inner sep=0,text=black,
          font=\fontfamily{phv}\fontsize{6.4}{7.4}\selectfont]
          at (0.695,0.0105) {Full search (full bar)};
      \end{scope}
    \end{tikzpicture}
    \caption{{\textbf{Fewer attempts to approach the other method's best score.}} {
      {Inner bars show research attempts needed to reach at least $90\%$ of the other method's best score; full bars show total attempts. \ourmodel reaches its reference score with fewer attempts on five of six tasks.}}}
    \label{fig:sample_efficiency}
    \vspace{-4pt}
\end{figure}
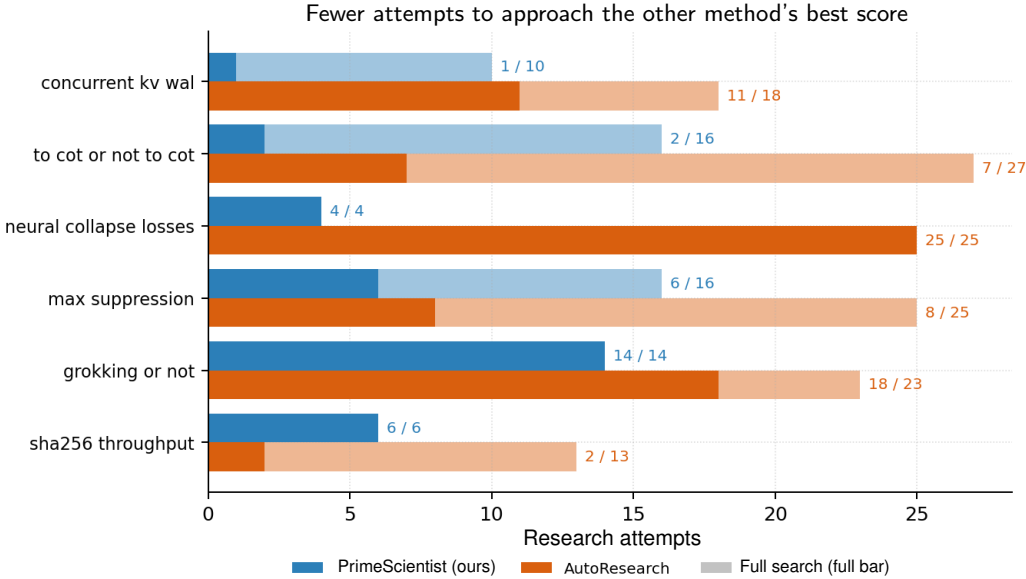

%% file: figures/fig_llm_value_consistency.tex
\begin{wrapfigure}{R}{0.52\textwidth}
  \vspace{-\intextsep}
  \centering
  \begin{tikzpicture}
    \node[anchor=south west,inner sep=0] (plot) at (0,0)
      {\includegraphics[width=\linewidth]{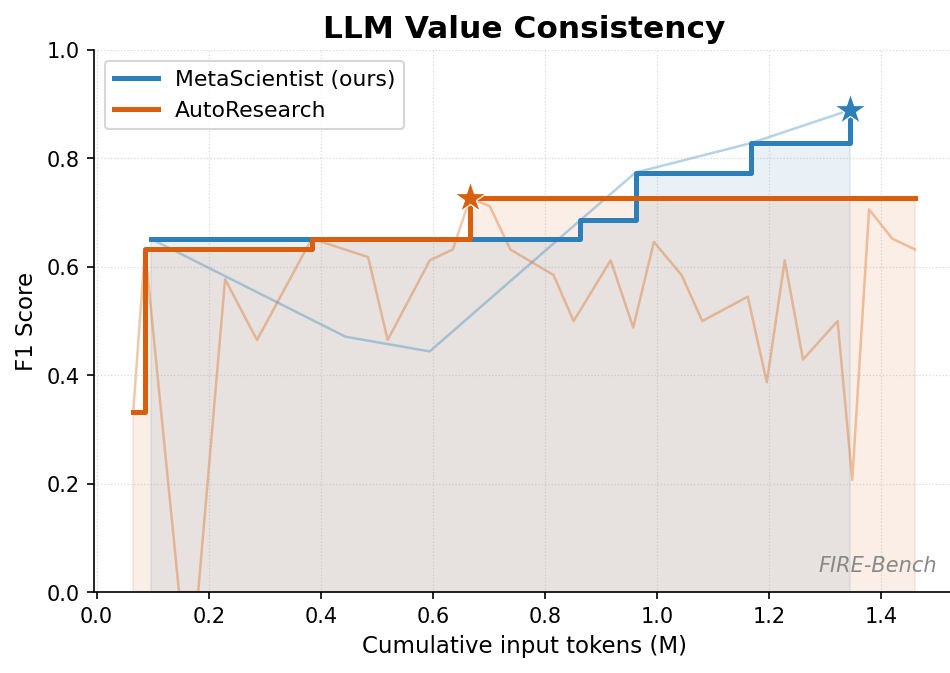}};
    \begin{scope}[x={(plot.south east)},y={(plot.north west)}]
      \fill[white] (0.178,0.864) rectangle (0.420,0.906);
      \node[anchor=base west,inner sep=0,text=black,
        font=\fontfamily{phv}\fontsize{5.3}{6.2}\selectfont]
        at (0.1842,0.8722) {\brandname{} (ours)};
    \end{scope}
  \end{tikzpicture}
  \captionsetup{skip=3pt,font=small,justification=raggedright,singlelinecheck=false}
  \caption{{\textbf{Research progress as input tokens accumulate.} On \tasknameinline{LLM Value Consistency}, \ourmodel continues improving after AutoResearch plateaus. Each method contributes one search; thin lines show per-attempt F1 and thick lines track the best score reached.}}
  \label{fig:llm_value_consistency}
  \vspace{-5pt}
\end{wrapfigure}

%% file: figures/fig_case_study_loa.tex
\begin{figure}[!t]
  \centering
  \includegraphics[width=\linewidth]{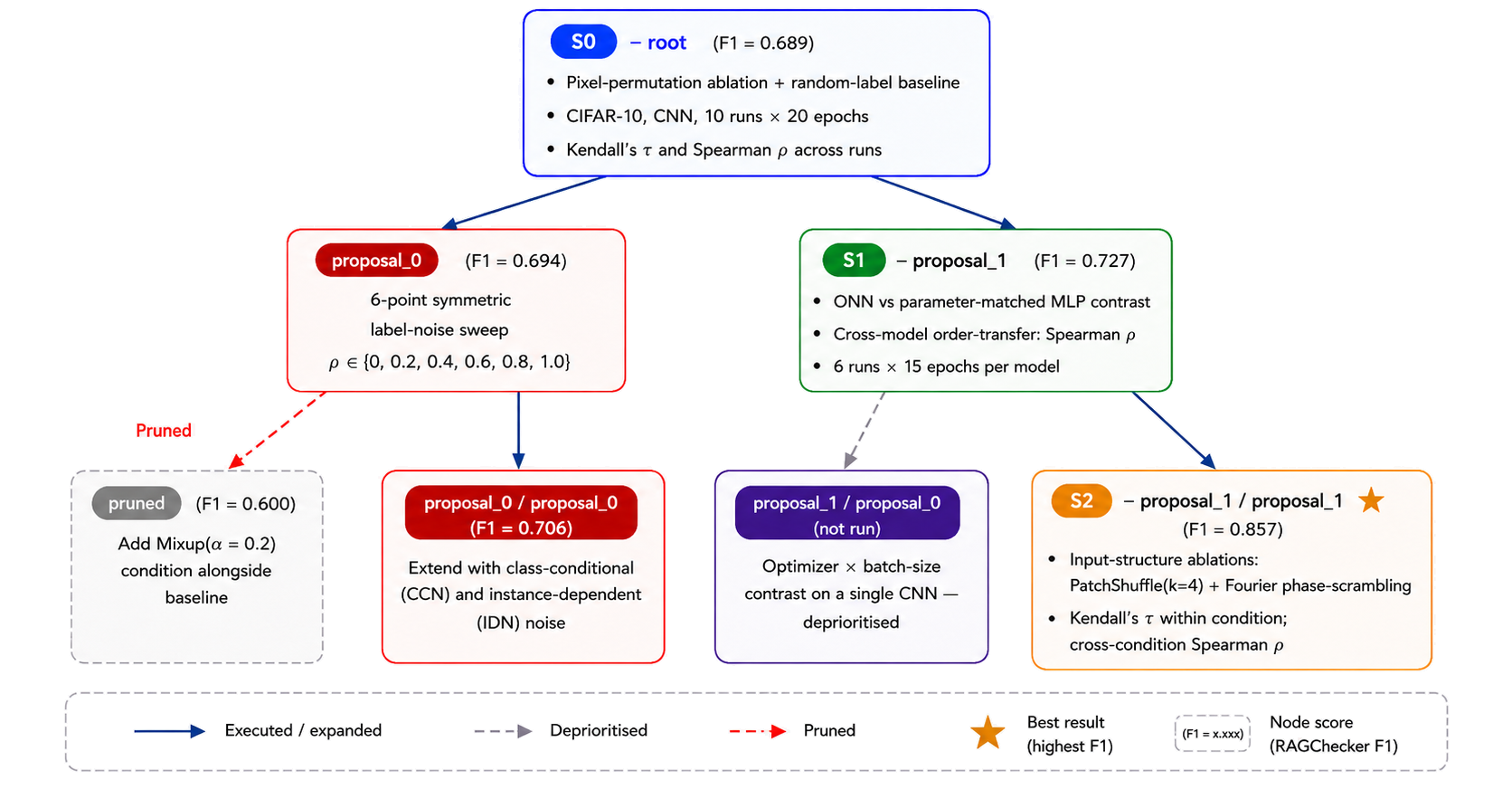}
  \caption{{\textbf{Selective plan expansion improves research findings.}
  On {\tasknameinline{Learning Order Agreement}}, \ourmodel studies why networks learn images in similar orders{~\citep{hacohen2020agree}}.
  {The tree develops label-noise and architecture experiments as alternative explanations. Image-structure tests in the architecture branch raise F1 from $0.689$ to $0.857$. F1 measures agreement with reference findings.}}}
  \label{fig:case_study_loa}
  \vspace{-4pt}
\end{figure}

%% file: sections/6_conclusion.tex
\Needspace{7\baselineskip}
\vspace{-6pt}
\section{Conclusion}
\label{sec:conclusion}
\vspace{-4pt}

{In this paper, we study strategic research effort allocation in autonomous research.}
{We introduce \ourmodel to make strategic effort allocation an explicit part of how autonomous agents conduct research.}
{To the best of our knowledge, we provide the first formulation that jointly models plan construction and complete research attempts under a shared inference {budget}.}
{\ourmodel's executable plan tree connects research attempts through retained alternatives and empirical outcomes, enabling adaptive MCTS-based allocation across research directions.}

\ourmodel achieves {better or comparable} research outcomes with {fewer research attempts than AutoResearch under the same resource budget}.
{Our ablation studies show that adapting exploration to the remaining budget improves overall reward over fixed exploration policies.}
Research-agent evaluation should therefore consider achieved outcomes, {research attempts}, and total inference costs jointly, including the resources consumed in deciding {which research direction to pursue next}.
{A longer-term direction is to learn allocation policies that improve how research agents use their resources, turning advances in AI and computation into scientific discovery at scale.}\looseness=-1
\par
\vspace{-7pt}

%% file: sections/7_appendix.tex
\section{Limitations}
\label{app:limitations}

\noindent \textbf{{Planning under limited resources.}}
{The reflector is a full LLM agent, and a reflection call can consume tokens on the same order as a coding-agent run. Sharing {budget} $B$ makes the balance between constructing alternatives and executing them explicit. The budget analysis in Appendix~\ref{app:budget_sensitivity} identifies task-dependent benefits at different {budgets}; characterizing the point at which further planning is preferable to immediate execution remains an open research direction.}

\noindent \textbf{{Scope of resource accounting.}}
{We measure inference cost in input and output tokens across both agents, and experimental effort in complete executor invocations. Each invocation constitutes one research attempt; its training workload, runtime, and monetary cost depend on the task and execution environment. A shared token {budget} constrains inference expenditure, while realized consumption depends on the completed actions and stopping point. Runtime is measured separately in Appendix~\ref{app:runtime}. Extending this accounting to monetary cost or energy requires the corresponding execution measurements, hardware characteristics, and inference pricing.}

\noindent \textbf{{Variability across independent searches.}}
{Complete research attempts provide task-level evidence for branch selection, with the amount of evidence determined by the available budget. On FIRE-Bench, each plan is executed twice and the higher score is retained; both invocations count toward inference consumption and experimental effort. The two or three independent full-search repetitions in the GPT-5-mini setting characterize variability through descriptive means and standard deviations. Larger repeated-search panels would enable more precise estimates of small reward differences. Statistical significance and backbone effects require separate analysis{. The primary GPT-5 setting uses} a \mbox{different {budget} and task coverage}.}

\noindent \textbf{Plan-level granularity.}
{\ourmodel} allocates complete research attempts and receives their empirical outcomes after execution. This granularity connects strategic decisions to task-level evidence, while leaving decisions within an attempt to the coding agent. Even a narrowly specified plan modification requires another complete attempt for evaluation. The current {method} therefore cannot reallocate effort among intermediate implementation decisions during an ongoing execution.

\section{Broader Impact}
\label{app:impact}

\noindent \textbf{{Accessible autonomous research.}}
Explicit allocation provides a {framework} for studying how limited inference resources should be distributed across research directions. {Achieving competitive outcomes with fewer complete research attempts can make autonomous research more accessible when empirical trial and error requires substantial implementation and experimentation. The framework includes planning in the inference {budget} and exposes the corresponding demand for complete research attempts, supporting resource-aware choices about research effort.} Application beyond the evaluated task classes also requires suitable execution environments and reliable feedback.

\noindent \textbf{{Strategic allocation for recursive self-improvement.}}
{A longer-term opportunity is \emph{recursive self-improvement}, in which research agents develop improved research procedures and use them to guide subsequent improvements~\citep{dgm,bilevel_autoresearch}. Our formulation offers a way to study how such systems divide limited resources between proposing changes to the researcher and empirically testing those changes. Strategic allocation could help sustain this improvement loop by directing effort toward promising changes while retaining alternatives for future evaluation.}

\noindent \textbf{{Scientific validation and reproducibility.}}
{Stored parent plans and executable modifications make research decisions traceable and support reconstruction of the plan tree. Reliable scientific use requires independent validation of conclusions beyond the scalar reward used to guide search. {Retaining execution records, documenting software environments, and using available held-out tests help verify reproducible improvements and detect evaluator exploitation.}}

\noindent \textbf{{Resource transparency.}}
{The reported inference {budgets} and numbers of research attempts document distinct uses of research resources. {End-to-end energy use depends on inference, experimental execution, and hardware utilization, requiring measurements beyond token and attempt counts.}}

\FloatBarrier
\section{Algorithm}
\label{app:algorithm}

\begin{algorithm}[H]
\caption{{\ourmodel Plan Search and Effort Allocation}}
\label{alg:bart}
\begin{algorithmic}[1]
\REQUIRE Task instruction $\mathcal{T}$, evaluator $h$, reflector $f$, selection policy $\pi$ (defined in \S\ref{sec:instantiation}), branching factor $m$, total agentic budget $B$, exponent cap $\alpha_{\max}$, prune threshold $\delta$
\STATE Initialise tree $T \leftarrow \{s_0\}$ where $s_0 \leftarrow f(\emptyset;\, \mathcal{T})$ \hfill\COMMENT{root executable plan drafted from instruction}
\STATE $U \leftarrow$ tokens consumed in drafting $s_0$
\WHILE{$U < B$}
    \STATE $\alpha_t \leftarrow \min\!\bigl(1/r_t,\; \alpha_{\max}\bigr),\; r_t \leftarrow (B-U)/B$
    \STATE $s \leftarrow \pi(T;\, \alpha_t)$ \hfill\COMMENT{research effort allocation; pruned nodes are skipped, so a node whose children are all pruned is returned as a leaf}
    \IF{$s$ unevaluated}
        {
        \STATE $\rho \leftarrow h(s)$; record the observed outcome at $s$
        \IF{$s\ne s_0$ and $\delta>0$ and $Q(\mathrm{parent}(s))>0$ and $\rho < Q(\mathrm{parent}(s)) - \delta$}
            \STATE exclude $s$ from subsequent selection \hfill\COMMENT{prune}
        \ELSE
            \STATE backpropagate $\rho$ to $s$ and its ancestors
        \ENDIF
        }
    \ELSE
        \STATE $\{s'_1,\dots,s'_m\} \leftarrow f\bigl(s;\,\Sigma(T)\bigr)$ \hfill\COMMENT{reflect; propose $m$ children}
        \STATE $T \leftarrow T \cup \{s'_1,\dots,s'_m\}$
    \ENDIF
    \STATE update $U$ with tokens consumed in this iteration
\ENDWHILE
\RETURN $\argmax_{s\in T:\,s\ \text{evaluated}} h(s)$
\end{algorithmic}
\end{algorithm}

{In Algorithm~\ref{alg:bart}, $\Sigma(T)$ denotes the tree context supplied to the reflector during plan expansion.}

\FloatBarrier
\input{sections/8_resource_appendix}
\FloatBarrier

\input{sections/9_task_settings}
\FloatBarrier

\Needspace{7\baselineskip}
\section{Skill Format}
\label{app:skill}

{Each research plan is a structured Markdown skill supplied to the coding agent before a research attempt.}
For FIRE-Bench, it contains: the research question decomposition, hypotheses to test, implementation plan, dataset handling strategy, and pitfalls to avoid.
For AutoLab, it contains: which files to edit, prioritized optimization techniques, correctness constraints, and expected performance targets.\looseness=-1

The root skill is written by the reflector from scratch given only the task instruction.
Each subsequent skill is produced by running the child's {\texttt{diff.py}}, which applies a focused modification to the parent's skill.
{The stored root plan and executable modifications allow each descendant plan to be reconstructed. Every modification records a reflector decision and is applied to its parent plan.}

\section{Reflector Prompt Template}
\label{app:reflector_prompt}

{During plan expansion, the reflector reads the selected node's execution outputs and proposes child plans.}
Listing~\ref{lst:reflector} is the full system prompt delivered to the reflector.
Runtime placeholders appear in \texttt{<angle-bracket>} form:
\texttt{<task-id>} is the benchmark task name,
\texttt{<node-path>} is the relative path to the completed node, and
\texttt{N} is the branching budget (maximum number of child proposals).
{The \texttt{[PRUNED NODES]} line appears only when earlier nodes have been pruned.}

\begin{lstlisting}[style=prompt,caption={Reflector prompt template with runtime placeholders in \texttt{<angle-bracket>} notation.},label={lst:reflector}]
You are a reflector for task: <task-id>

A new run has just completed. Your working directory is the search tree
root for this task.  The completed node is at: <node-path>/

If this is not your first call on this task, you are RESUMING a previous
session -- you already have memory of files and proposals you inspected
last time.  Do NOT re-read files whose content you already remember; just
check for NEW files (e.g. the newly-completed node's packet.json, result
directories) and whatever sibling state has changed since your last call.

Relevant files (anything not listed here is noise -- skip it):
  [PRUNED NODES]
  <node-path>/packet.json     -- reward score (precision/recall/f1) and run
                                 metadata.  Reflects correctness of the
                                 agent's conclusion.
  <node-path>/log_brief.log   -- head+tail excerpt of the agent run.  Primary
                                 diagnostic tool: look for Python tracebacks,
                                 missing-file errors, or the agent admitting
                                 it could not finish.  The agent often claims
                                 success -- trust packet.json, not the narrative.
  <node-path>/.log.log        -- FULL agent log (very large).  Read ONLY if
                                 log_brief.log leaves the failure ambiguous
                                 (e.g. tail ends mid-traceback).
                                 Do NOT read routinely; use only when the
                                 reward score is very low.
  <node-path>/skill.md        -- the experimental plan this run executed.
  <node-path>/changes.md      -- (if present) short prose summary of what
                                 this node changed relative to its parent.
  <node-path>/sandbox/        -- the agent's experiment workspace.  Use `ls`
                                 to discover result directories, then read
                                 specific output files only if needed.
  Other nodes' prior.json / skill.md / packet.json -- to review previous
                                 trials' approaches and performance.

Your job:
1. Read packet.json for the score.  Read log_brief.log for the run
   trajectory.  Use `ls <node-path>/sandbox/` to find result directories,
   then read specific files if the log is insufficient.
   Ground your analysis in concrete observations, not assumptions.

2. Check other nodes' prior.json / skill.md / packet.json as needed to
   avoid re-proposing already-tried hypotheses and to learn what has and
   has not worked.

3. Identify whether failure was due to a setup/runtime error (fixable by
   changing the experimental plan) or a genuine result (hypothesis tested
   but scored low).

4. Propose between 1 and N CONTROVERSIALLY DIFFERENT skill variants.
   "Controversially different" means each proposal bets on a fundamentally
   different hypothesis -- different hyperparameter regime, different metric
   interpretation, different dataset choices, etc.  Do NOT create proposals
   that differ only in a minor hyperparameter value or wording; those are
   not separate bets.

   Decide how many to create:
   - One clearly dominant direction: write 1 proposal.
   - Two to N genuinely competing hypotheses: write that many.
   - Never pad with minor variations just to hit a count.

   Create child directories numbered from where you left off:
     <node-path>/children/proposal_0/
     <node-path>/children/proposal_1/
     ...  (up to proposal_{N-1}/)

   In each child directory write THREE files (and run the diff):

   a) diff.py -- a self-contained Python script that reads the parent's
      skill.md and writes a NEW COMPLETE skill.md into this directory.
      The output skill.md must be a FULL self-contained experimental plan
      (dataset, model, hyperparameters, evaluation metrics, conclusion
      structure) -- NOT a delta.  Template:

        from pathlib import Path
        parent = Path(__file__).parents[2] / "skill.md"
        output = Path(__file__).parent     / "skill.md"
        text   = parent.read_text()
        # Apply targeted transformations, e.g.:
        # text = text.replace("model = X", "model = Y")
        output.write_text(text)

   b) Run it immediately:  python diff.py
      Verify skill.md was created and is the full updated plan.
      If diff.py errors, fix it and re-run before continuing.

   c) changes.md  (< 1 KB) -- short prose summary written FOR the child
      agent (3-6 bullet points), e.g.:
        - switched dataset from X to Y
        - moved evaluation metric from accuracy to F1
        - added ablation on hyperparameter Z
      The child agent reads this in its inherited sandbox to know what
      to update without repeating the parent's work.

   d) prior.json -- structured reasoning for the next reflector to consult:
        {
          "estimate":   <float 0-1>,
          "hypothesis": "<single concrete hypothesis this proposal tests>",
          "rationale":  "<2-3 sentences: WHY this should improve the score,
                          citing observations from packet.json / log_brief.log
                          / sandbox>",
          "changes":    "<1 sentence: which files/sections are modified and
                          at what abstraction level>",
          "risks":      "<1-2 sentences: most plausible failure mode>"
        }

   Do NOT write skill.md directly -- only via diff.py.

Print a one-line summary for each proposal and each pruned branch.
\end{lstlisting}

\section{Example Research Plan and Modification}
\label{app:example_plan}

\subsection{Original Plan}

\begin{lstlisting}[style=prompt,caption={Reflector-generated root plan (\texttt{skill.md}) for the activation-control task.},label={lst:original_plan}]
# Activation Control: Experimental Plan and Reproduction Guide

This file specifies an exact, 1-hour-per-run plan to investigate:

Can we efficiently elicit long chain-of-thought reasoning in language models through activation-level interventions?

It defines fixed datasets, models, metrics, hyperparameters, and a step-by-step procedure. No outcomes are assumed; record all actual results to `results.tsv`.

## 1) Datasets (fixed)

- GSM8K (grade school math word problems)
  - Loader: `load_dataset("openai/gsm8k", "main")`
  - Splits:
    - Calibration: `train[:64]` (64 items)
    - Evaluation: `test[:100]` (100 items)
  - Answer format: final numeric answer; evaluate by exact numeric match after normalization (strip, remove commas, allow leading/trailing spaces, allow enclosing `\boxed{...}`); ignore units.

- MMLU (abstract_algebra)
  - Loader: `load_dataset("cais/mmlu", "abstract_algebra")`
  - Splits:
    - Calibration: `validation[:64]` (64 items)
    - Evaluation: `validation[64:164]` (100 items)
  - Format: multiple choice with options A/B/C/D; evaluate by exact choice match.

These sample sizes are chosen to keep each run about 1 hour on a single 7B model with moderate batch sizes and `max_new_tokens<=256`.

## 2) Models (fixed)

Load via HuggingFace through the provided helper:

- `Qwen/Qwen2.5-7B`
- `Qwen/Qwen2.5-7B-Instruct`
- `Qwen/Qwen2.5-Math-7B`

Use `from utils.llm_inference import LLMInference` and its `batch_generate()` for batched decoding. Unless otherwise noted, run with the Instruct variant for GSM8K and the Math variant for MMLU; also include the Base model to test generality.

## 3) Decoding and runtime (fixed)

- `temperature`: 0.2 (stable reasoning); also probe 0.7 in calibration only when building steering vectors (see Section 5.2), not in evaluation.
- `top_p`: 0.95
- `max_new_tokens`: 256
- `stop`: none (allow natural stop or EOS)
- `batch_size`: 8 (tune to memory, but keep >=4; record the actual value used)
- `seed`: 1234
- `repetition_penalty`: 1.0
- Determinism: set torch no-grad and eval mode; disable dropout if applicable.

## 4) Metrics (fixed)

- Reasoning length: number of generated tokens per example (counted over the entire assistant completion). Report mean and median per condition.
- Final-answer correctness:
  - GSM8K: extract last number using regex `[\-\+]?\d+(?:,?\d)*(?:\.\d+)?` from the final line or from within `\boxed{}` if present; exact match to gold after removing commas.
  - MMLU: exact letter match among {A,B,C,D}.
- Throughput: tokens/sec (optional), for budget awareness.

## 5) Conditions and interventions

Always include prompting baselines, then add activation-level methods. Run the same eval subsets (Section 1) for every condition.

### 5.1 Prompting baselines (no activation edits)

- `direct`: Short answer only. System/user prompt ends with: "Give only the final answer."
- `cot`: Chain-of-thought. Append: "Let's think step by step." (no activation edits).

### 5.2 Activation-level methods

Implement using forward hooks over the model's residual stream. Let the model have L transformer blocks. Define three layer bands by index (0-based, inclusive):

- Early:    range(round(0.20*L), round(0.30*L))
- Middle:   range(round(0.45*L), round(0.60*L))  (default)
- Late:     range(round(0.75*L), round(0.85*L))

Apply per-token, per-layer additive interventions to the hidden state h immediately after the block output (post-attn+MLP, pre-residual add), implemented as h := h + alpha * v, where v is a cached steering vector (same dimension as h). Use FP16/FP32 matching the model dtype.

Steering vectors are computed on calibration sets (Section 1) using hidden states collected with `output_hidden_states=True` and the following prompts:

- Long-thinking prefix: "Let's think step by step."
- Short-answer prefix:  "Answer concisely."

Compute mean hidden states over the prefix tokens only.

Methods:

- `SV+ (long)`: Single-vector steering toward long thinking.
  - v = mean(h | long-thinking) - mean(h | neutral). Neutral uses no extra instruction beyond task template.
  - Hyperparameters: alpha in {0.5, 1.0, 2.0, 3.0}; layers in {Early, Middle, Late}.

- `CAS (contrastive)`: Contrastive activation steering.
  - v = mean(h | long-thinking) - mean(h | short-answer).
  - Hyperparameters: alpha in {0.5, 1.0, 2.0, 3.0}; layers in {Early, Middle, Late}.

- `Patch-N`: Activation patching for the first N generation steps.
  - Record hidden states from a run with the long-thinking prefix on calibration items; at evaluation, for each test prompt, replace the layer outputs in the chosen band with the recorded mean over calibration for the first N steps.
  - Hyperparameters: N in {8, 16, 32}; layers in {Middle} only (to control budget).

Notes:
- Build vectors separately per model (not shared across models).
- Cache one vector per method x layer-band in artifacts/{model}/steering/{method}-{band}.pt.

## 6) Experimental matrix (fixed)

For each model in Section 2, run the following conditions on each evaluation subset in Section 1:

- direct
- cot
- SV+ (alpha in {0.5,1,2,3}, band in {Early,Middle,Late})
- CAS  (alpha in {0.5,1,2,3}, band in {Early,Middle,Late})
- Patch-N (N in {8,16,32}, band=Middle)

This yields 2 + (4x3) + (4x3) + 3 = 29 conditions per model. To keep runs within about 1 hour, use batch_generate() and process datasets in batches; if time is tight, prioritize Middle band first, then Early/Late.

## 7) Implementation guide (step-by-step)

The codebase provides utils.llm_inference.LLMInference with batch_generate(). Implement only light wrappers and hooks.

1. Environment
   - Install: pip install datasets transformers accelerate einops (and flash-attn if available).
   - Confirm GPU if available; otherwise reduce batch_size to fit RAM.

2. Loader
   - Write experiments/datasets.py with two helpers: load_gsm8k(calib_size=64, eval_size=100) and load_mmlu_aa(calib_size=64, eval_size=100) that return (calib, eval) lists of dicts with fields: id, prompt, answer (gold), and an extract_fn callable for scoring.

3. Prompts
   - GSM8K template: "Solve the math problem. {question}\n"
   - MMLU template: "Choose the correct option (A/B/C/D). {question}\nOptions:\nA) ... B) ... C) ... D) ...\nAnswer with a single letter."
   - Baseline suffixes: direct -> "Give only the final answer."; cot -> "Let's think step by step."

4. Hidden-state capture
   - In a module experiments/steering.py, implement:
     - collect_prefix_hidden_states(model, tokenizer, prompts, layers_band) -> tensor [num_layers, d_model] averaged over prefix tokens; use output_hidden_states=True and register forward hooks on the target blocks to read their outputs.
     - build_vector(method, long_prompts, short_prompts, neutral_prompts) that returns a dict {band: vector}.
     - apply_steering_hooks(model, vector, alpha, layers_band) that adds an in-place forward hook performing h += alpha*vector[layer_idx] at the chosen layers during generation.
     - apply_patching_hooks(model, cached_states, N, layers_band) that, for generation steps < N, replaces h with the cached mean state for that time step.

5. Calibration (per model)
   - Use calibration splits (Section 1) to construct prompts and compute SV+ and CAS vectors and to record patching states.
   - Use temperature=0.7 during the long-thinking runs when collecting states (encourages richer trajectories); evaluation always uses Section 3 decoding.
   - Save vectors to artifacts/{model}/steering/ and patch caches to artifacts/{model}/patch/.

6. Evaluation loop
   - For each condition in Section 6, attach the appropriate hooks (or none for baselines) and call batch_generate() on the evaluation items with decoding settings in Section 3.
   - For each output, compute:
     - gen_len_tokens
     - is_correct via the dataset-specific extract_fn
   - Append one TSV row per condition with the schema below.

## 8) Logging format (fixed)

Append to results.tsv using tab-separated columns (one header row if file is empty):

model	dataset	condition	alpha	band	N	seed	temperature	max_new_tokens	batch_size	n_items	acc	mean_len	median_len

- Use alpha for SV+/CAS (empty for others), N for Patch-N (empty for others), and band in {Early,Middle,Late} or empty for baselines.
- acc is fraction in [0,1]. mean_len/median_len are in tokens.

## 9) Correctness constraints

- Do not leak gold answers into calibration prompts.
- Use calibration items only to build vectors and set hyperparameters; do not include them in evaluation metrics.
- Keep all other settings identical across conditions (Section 3) to isolate activation effects.
- Ensure hooks are removed/reset between conditions.
- For GSM8K, parse only the final numeric answer; ignore intermediate reasoning content.
- For MMLU, force output to the set {A,B,C,D}; if the generation contains more text, extract the first valid letter.

## 10) Order of execution (to fit about 1 hour)

For each model (start with Qwen2.5-7B-Instruct):

1) Build vectors/caches (calibration, Section 5) for Middle band only.
2) Evaluate: direct, cot, SV+ (alpha in {0.5,1,2,3}, band=Middle), CAS (alpha in {0.5,1,2,3}, band=Middle), Patch-N (N in {8,16,32}, band=Middle).
3) If time remains, add Early then Late bands for SV+/CAS.
4) Repeat for the Base and Math variants on the dataset most suited to them (Base on GSM8K, Math on MMLU first), then cross-evaluate if time allows.

## 11) Reproducibility

- Record the exact package versions and GPU/CPU info at the top of results.tsv as commented lines starting with #.
- Save all built vectors and caches under artifacts/ and include a JSON manifest describing layer indices and shapes.

## 12) Quick-start checklist

- [ ] Install deps (Section 7.1)
- [ ] Implement minimal hooks (Section 7.4)
- [ ] Build vectors on calibration (Section 7.5)
- [ ] Run evaluation matrix (Section 7.6) with Section 10 order
- [ ] Append rows to results.tsv using Section 8 schema
- [ ] Summarize trends (post-hoc; do not assume outcomes in advance)
\end{lstlisting}

\Needspace{10\baselineskip}
\subsection{Modification}

\begin{lstlisting}[style=pyplan,caption={Reflector-generated \texttt{diff.py} for a projection-based intervention.},label={lst:modification}]
from pathlib import Path

parent = Path(__file__).parents[2] / "skill.md"
output = Path(__file__).parent / "skill.md"

text = parent.read_text()

# Identify the start of Section 5.2 and the start of Section 6 to replace the activation
# methods and the experimental matrix with a new, self-contained variant.
start_marker = "### 5.2 Activation-level methods"
end_marker = "## 6) Experimental matrix"

if start_marker in text and end_marker in text:
    pre = text.split(start_marker)[0]
    post = text.split(end_marker, 1)[1]
else:
    # Fallback: write the original text verbatim (should not happen), then append our variant.
    pre = text
    post = ""

new_52 = f"""{start_marker}

Replace additive steering and patching with a projection-based, channel-sparse, time-gated method that suppresses the
"concise-answer" subspace rather than pushing toward a fixed long-CoT vector. This aims to lengthen reasoning while
minimizing off-manifold drift.

Methods (projection family):

- OP-Short (Orthogonal Projection against Shortness):
  - Learn a linear probe w on hidden states to discriminate between prompts with a short-answer suffix vs. no suffix.
    Use calibration prefixes only. Fit a logistic regression on pooled hidden states from the target layer band.
  - Let u = w / ||w|| be the unit "shortness" direction. At generation, apply h := h - β (h·u) u (orthogonal projection
    removing the shortness component). This suppresses concise-answer bias without forcing a particular long vector.
  - Channel-sparse variant: compute the top-k channels by |u| and apply projection only over those dimensions; k ∈ {64, 128}.

- OP-Short+Gate (Time-gated):
  - Apply OP-Short only for the first K decoding steps, with K ∈ {16, 32}. This concentrates intervention where planning
    tokens occur, reducing late-stage derailment.

Layer bands:
- Middle (default), Late (secondary). Early is excluded to save budget and avoid destabilizing tokenization dynamics.

Hyperparameters:
- β ∈ {0.25, 0.5, 1.0}
- k ∈ {64, 128}
- K ∈ {16, 32}

Implementation notes:
- Hidden-state collection: use output_hidden_states=True and pool prefix tokens only (excluding the question body when possible).
- Probe training: L2-regularized logistic regression on CPU (scikit-learn or a simple torch optimizer over a single layer).
- Hook: register a forward hook after the block output in the chosen band; implement h := h - β (h·u) u with optional
  masking to the top-k channels. Ensure dtype consistency (FP16/FP32) and remove hooks between conditions.
- Caching: save u and channel indices per model×band under artifacts/{{model}}/opshort/.
"""

new_6 = """## 6) Experimental matrix (fixed)

For each model in Section 2 and each evaluation subset in Section 1, run:

- direct (short answer only)
- cot ("Let's think step by step.")
- OP-Short (β ∈ {0.25, 0.5, 1.0}; band ∈ {Middle, Late}; k ∈ {64,128})
- OP-Short+Gate (same as OP-Short with K ∈ {16, 32})

This replaces SV+/CAS/Patch-N from the parent plan. Keep Section 3 decoding fixed for comparability. Prioritize the Middle band
first to fit within time; evaluate Late band if time remains.
"""

# Optional note on lightweight fallback models to ensure runnable evidence when GPU is unavailable.
fallback_note = """

Addendum (lightweight fallback):
- If 7B models are infeasible on available hardware, include a smaller HF model such as TinyLlama-1.1B-Chat as an auxiliary
  target to demonstrate the length-vs-accuracy tradeoff under OP-Short. Clearly mark such runs as exploratory and keep the
  primary conclusions tied to Qwen 7B models when resources permit.
"""

new_text = pre + new_52 + "\n\n" + new_6 + fallback_note

# Ensure the resulting plan remains self-contained by keeping the remainder of the parent doc (Sections 7+).
if post:
    # Reattach the rest of the document starting from the original Section 6 header content forward, but our Section 6
    # has been replaced; therefore, we need to find the next section after 6), which is "## 7) Implementation guide".
    # Split post at the first occurrence of "## 7)" if present.
    marker7 = "## 7) Implementation guide"
    if marker7 in post:
        after6 = post.split(marker7, 1)[1]
        new_text += "\n\n" + marker7 + after6
    else:
        # If not found, just append the remainder to avoid losing content.
        new_text += post

output.write_text(new_text)
print(f"Wrote updated skill.md to {output}")
\end{lstlisting}

%% file: sections/8_resource_appendix.tex
\par\begingroup
\setlength{\intextsep}{10pt plus 2pt minus 2pt}
\section{{Resource Use and Budget Sensitivity}}
\label{app:resource_analysis}

{The main results show that \ourmodel can achieve comparable or better research outcomes with fewer attempts. The AutoLab analyses below examine how this benefit relates to resource use. We first account for the inference spent on planning and execution, then compare research quality at matched token budgets and attempt counts. We next examine how outcomes change with the total budget and measure elapsed search time. These analyses provide complementary views of research efficiency, with the comparison conditions specified separately for each study.}

\subsection{{Inference Allocation and Experimental Effort}}
\label{app:resource_accounting}

{Planning consumes inference resources before a research direction is tested. To understand this expenditure, we group token use into coding-agent execution and planning, including reflection, tree context, and child-plan generation (Table~\ref{tab:token_allocation}). This accounting captures the resources used to choose research directions alongside those used to execute them.}

\begin{table}[!ht]
\centering
\caption{{\textbf{Inference spent on planning and execution.} Planning accounts for a substantial share of token use, including reflection, tree context, and child-plan generation. Both components contribute to the total inference expenditure.}}
\label{tab:token_allocation}
\small
\setlength{\tabcolsep}{10pt}
\begin{tabular}{@{}lr@{}}
\toprule
\textbf{Component} & \textbf{{Token share}} \\
\midrule
Coding-agent tokens & $62\%$ \\
Reflection, tree context, and child-plan generation & $38\%$ \\
\bottomrule
\end{tabular}
\end{table}

{Coding-agent execution accounts for $62\%$ of the token use in Table~\ref{tab:token_allocation}, while planning accounts for $38\%$. The reflector alone uses $21\%$ to $37\%$ of tokens per run, and a plan expansion consumes $0.1$ to $0.3$ million tokens. Planning is therefore a substantial part of the shared budget. The next comparison examines the number of research attempts performed after including this planning expenditure.}

{For the six-task comparison in Table~\ref{tab:attempt_counts}, each method receives a total budget of 1M tokens per task. \ourmodel performs $112$ research attempts across the six tasks, compared with $158$ for AutoResearch, a reduction of approximately $29\%$. It uses fewer attempts on every task; on {\tasknameinline{Hash~Join}}, for example, the count falls from $34$ to $21$. Section~\ref{sec:reproducibility} examines consistency across repeated searches, while the following subsection examines the quality of the resulting solutions.}

\begin{table}[!ht]
\centering
\caption{\textbf{Fewer research attempts under a fixed inference budget.} {\ourmodel uses approximately $29\%$ fewer attempts across six AutoLab tasks ($112$ versus $158$). Each method has 1M tokens per task for planning and execution.}}
\label{tab:attempt_counts}
\small
\setlength{\tabcolsep}{14pt}
\begin{tabular}{@{}lrr@{}}
\toprule
\textbf{Task} & \textbf{\ourmodel} & \textbf{AutoResearch} \\
\midrule
{\taskname{Gaussian Blur}} & 18 & 22 \\
{\taskname{Hash Join}} & 21 & 34 \\
{\taskname{FFT (Rust)}} & 30 & 42 \\
{\taskname{Concurrent KV WAL}} & 11 & 24 \\
{\taskname{Flash Attention}} & 18 & 19 \\
{\taskname{Sha256 Throughput}} & 14 & 17 \\
\midrule
\textbf{Total} & \textbf{112} & \textbf{158} \\
\bottomrule
\end{tabular}
\end{table}

\FloatBarrier
\subsection{{Reward Under Matched Resources}}
\label{app:matched_resources}

{To assess the quality obtained from the available resources, we compare \ourmodel and AutoResearch under two matching criteria. The first fixes total inference, including both planning and execution. The second fixes the number of research attempts, giving both methods the same number of empirical tests within each task.}

{At the shared checkpoint of 1M tokens, Table~\ref{tab:token_checkpoint} summarizes rewards over three searches per method and task. The rewards are comparable across the six tasks, with \ourmodel reaching $0.583$ versus $0.561$ on {\tasknameinline{Concurrent KV WAL}} and $0.141$ versus $0.130$ on {\tasknameinline{Sha256 Throughput}}. These results show that \ourmodel preserves competitive solution quality when its planning and execution must share the same total inference budget.}

\begin{table}[!ht]
\centering
\caption{{\textbf{Research quality at matched token budgets.} AutoLab rewards remain comparable when each method has 1M tokens for planning and execution. Values are mean $\pm$ standard deviation across three searches per method and task.}}
\label{tab:token_checkpoint}
\small
\setlength{\tabcolsep}{12pt}
\begin{tabular}{@{}lcc@{}}
\toprule
\textbf{Task} & \textbf{\ourmodel} & \textbf{AutoResearch} \\
\midrule
{\taskname{Gaussian Blur}} & $0.460 \pm 0.03$ & $0.485 \pm 0.02$ \\
{\taskname{Hash Join}} & $0.626 \pm 0.00$ & $0.629 \pm 0.01$ \\
{\taskname{FFT (Rust)}} & $0.499 \pm 0.01$ & $0.513 \pm 0.02$ \\
{\taskname{Concurrent KV WAL}} & $0.583 \pm 0.02$ & $0.561 \pm 0.04$ \\
{\taskname{Flash Attention}} & $0.294 \pm 0.04$ & $0.296 \pm 0.01$ \\
{\taskname{Sha256 Throughput}} & $0.141 \pm 0.15$ & $0.130 \pm 0.15$ \\
\bottomrule
\end{tabular}
\end{table}

{We next match the number of research attempts within each task to examine the quality achieved with the same amount of experimentation. Table~\ref{tab:matched_attempts} presents the three improved cases from a six-task comparison. \ourmodel achieves higher rewards on {\tasknameinline{Hash Join}}, {\tasknameinline{FFT (Rust)}}, and {\tasknameinline{Concurrent KV WAL}}; on the latter, reward increases from $0.551$ to $0.583$. These cases show that strategic allocation can produce stronger outcomes from the same number of research attempts.}

\begin{table}[!ht]
\centering
\caption{{\textbf{Higher reward at matched research attempts.} The table shows the three improved cases from a six-task AutoLab comparison with the number of research attempts matched within each task.}}
\label{tab:matched_attempts}
\small
\setlength{\tabcolsep}{14pt}
\begin{tabular}{@{}lcc@{}}
\toprule
\textbf{Task} & \textbf{\ourmodel} & \textbf{AutoResearch} \\
\midrule
{\taskname{Hash Join}} & $0.625$ & $0.617$ \\
{\taskname{FFT (Rust)}} & $0.498$ & $0.493$ \\
{\taskname{Concurrent KV WAL}} & $0.583$ & $0.551$ \\
\bottomrule
\end{tabular}
\end{table}

\FloatBarrier
\Needspace{7\baselineskip}
\subsection{{Sensitivity to the Token Budget}}
\label{app:budget_sensitivity}

{A larger budget creates room both to construct alternative plans and to execute them. The benefit depends on whether the additional research effort leads to stronger solutions. We compare \ourmodel and AutoResearch on three AutoLab tasks at four total token budgets, from 250k to 1M, with the same budget assigned to both methods in each comparison (Table~\ref{tab:budget_sensitivity}). This sweep examines when additional resources improve outcomes and how the benefit varies across tasks.}

{{\tasknameinline{Concurrent KV WAL}} shows the clearest benefit from additional resources for \ourmodel. Its reward rises from $0.42$ to $0.57$ as the budget grows from 250k to 1M tokens, and it leads at 750k and 1M tokens. On {\tasknameinline{Sha256 Throughput}}, both methods improve substantially between 500k and 750k tokens, reaching $0.19$ at 750k. On {\tasknameinline{Flash Attention}}, rewards range from $0.26$ to $0.33$ across the two methods and four budgets. The sweep thus distinguishes gains shared by both methods from settings where strategic allocation produces stronger outcomes.}

\begin{table}[!ht]
\centering
\caption{{\textbf{Task-dependent gains from additional research resources.} The sweep compares rewards at four total token budgets; entries list \ourmodel{} / AutoResearch. At 750k and 1M tokens, \ourmodel achieves higher reward on \tasknameinline{Concurrent KV WAL}.}}
\label{tab:budget_sensitivity}
\small
\setlength{\tabcolsep}{7pt}
\begin{tabular}{@{}lcccc@{}}
\toprule
\textbf{Task} & \textbf{250k} & \textbf{500k} & \textbf{750k} & \textbf{1M} \\
\midrule
{\taskname{Concurrent KV WAL}} & $0.42 / 0.52$ & $0.47 / 0.51$ & $0.52 / 0.49$ & $0.57 / 0.54$ \\
{\taskname{Flash Attention}} & $0.30 / 0.30$ & $0.26 / 0.33$ & $0.30 / 0.32$ & $0.27 / 0.30$ \\
{\taskname{Sha256 Throughput}} & $0.02 / 0.02$ & $0.01 / 0.03$ & $0.19 / 0.19$ & $0.20 / 0.18$ \\
\bottomrule
\end{tabular}
\end{table}

\FloatBarrier
\Needspace{7\baselineskip}
\subsection{{Elapsed Search Time}}
\label{app:runtime}

{Elapsed time measures how long a user waits for the research outcome. Work on agent infrastructure shows that inference serving and workflow scheduling also shape this duration~\citep{pancake,flashevolve}. We therefore measure search time directly, complementing inference tokens and research attempts with the time required to deliver a result. Table~\ref{tab:runtime} reports durations across complete searches.}

{The comparison includes twelve AutoLab searches per method. \ourmodel completes these searches in $32$ minutes on average, compared with $42$ minutes for AutoResearch; the median is $34$ minutes for both. The lower average completion time provides an additional measure of practical efficiency alongside inference tokens and research attempts.}

\begin{table}[!ht]
\centering
\caption{{\textbf{Reduced elapsed time for AutoLab searches.} In 12 searches per method, mean time falls from 42 to 32 minutes with \ourmodel; the median is 34 minutes for both methods.}}
\label{tab:runtime}
\small
\setlength{\tabcolsep}{12pt}
\begin{tabular}{@{}lrrr@{}}
\toprule
\textbf{Method} & \textbf{Runs} & \textbf{Mean (min)} & \textbf{Median (min)} \\
\midrule
\ourmodel & 12 & 32 & 34 \\
AutoResearch & 12 & 42 & 34 \\
\bottomrule
\end{tabular}
\end{table}

\par\endgroup

%% file: sections/9_task_settings.tex
\Needspace{7\baselineskip}
\section{{Benchmark Tasks and Evaluation Targets}}
\label{app:research_tasks}

{The benchmarks cover distinct research objectives. FIRE-Bench tasks study AI research questions; AutoLab tasks require faster or smaller implementations; MLE-Bench tasks require predictive models for competition datasets. Tables~\ref{tab:fire_task_settings}, \ref{tab:autolab_task_guide}, and~\ref{tab:mle_task_guide} explain the questions, optimization goals, and evaluation targets behind the task names. The agent develops and tests its plans under the experimental budgets in Section~\ref{sec:setup}.}

\begin{table}[!ht]
\centering
\caption{{\textbf{Research questions behind the FIRE-Bench tasks.} Each row explains the phenomenon studied and identifies its source paper. The agent designs and conducts experiments to answer each question; F1 measures agreement between its conclusions and the reference findings.}}
\label{tab:fire_task_settings}
\small
\setlength{\tabcolsep}{5pt}
\renewcommand{\arraystretch}{1.13}
\begin{tabular}{@{}>{\raggedright\arraybackslash}p{0.29\textwidth}>{\raggedright\arraybackslash}p{\dimexpr0.71\textwidth-10pt\relax}@{}}
\toprule
{\normalcolor \textbf{Task and source}} & {\normalcolor \textbf{What the agent studies}} \\
\midrule
{\normalcolor {\normalcolor \taskname{Activation Control}}~\citep{zhao2025activation}} & {\normalcolor Intervene on internal LLM activations during inference and examine changes in reasoning length, self-reflection, and answer accuracy.} \\
\addlinespace[4pt]
{\normalcolor {\normalcolor \taskname{SECA Hallucination}}~\citep{liang2025seca}} & {\normalcolor Rephrase questions while preserving their meaning and coherence to test whether plausible wording changes induce hallucinations.} \\
\addlinespace[4pt]
{\normalcolor {\normalcolor \taskname{QuestBench}}~\citep{li2025questbench}} & {\normalcolor Present reasoning problems with missing information and test whether an LLM selects the clarification question needed to solve them.} \\
\addlinespace[4pt]
{\normalcolor {\normalcolor \taskname{Learning Order Agreement}}~\citep{hacohen2020agree}} & {\normalcolor Track which examples neural networks learn to classify earlier during training. Compare independently trained models and altered datasets to explain similarities in this order.} \\
\addlinespace[4pt]
{\normalcolor {\normalcolor \taskname{LLM Value Consistency}}~\citep{rozen2025values}} & {\normalcolor Elicit responses to human-value questionnaires under different contexts and framings. Examine whether the expressed values form a consistent structure, including correlations between values.} \\
\addlinespace[4pt]
{\normalcolor {\normalcolor \taskname{To CoT or not to CoT}}~\citep{sprague2025tocot}} & {\normalcolor Compare direct answers with chain-of-thought reasoning across task types to identify when intermediate reasoning improves accuracy.} \\
\addlinespace[4pt]
{\normalcolor {\normalcolor \taskname{Grokking or Not}}~\citep{doshi2024grok}} & {\normalcolor Train networks on modular arithmetic with corrupted labels to study how memorization, generalization, and regularization interact during training.} \\
\addlinespace[4pt]
{\normalcolor {\normalcolor \taskname{Max Suppression}}~\citep{zhou2025maxsup}} & {\normalcolor Compare label smoothing with maximum-logit suppression in image classification, examining prediction confidence and variation among features from the same class.} \\
\addlinespace[4pt]
{\normalcolor {\normalcolor \taskname{Neural Collapse Losses}}~\citep{zhou2022neuralcollapse}} & {\normalcolor Train classifiers with different losses and examine whether their learned features converge to similar within-class clusters and between-class geometry.} \\
\addlinespace[4pt]
{\normalcolor {\normalcolor \taskname{CoT Faithfulness Gaps}}~\citep{chen2025faithfulness}} & {\normalcolor Insert answer hints into reasoning questions and examine whether a model acknowledges the hints when they influence its answer.} \\
\addlinespace[4pt]
{\normalcolor {\normalcolor \taskname{Lost in The Middle}}~\citep{liu2024lost}} & {\normalcolor Vary the position of relevant information in long input contexts and measure its effect on question answering and retrieval performance.} \\
\addlinespace[4pt]
{\normalcolor {\normalcolor \taskname{MCQ Selection Bias}}~\citep{zheng2024selectors}} & {\normalcolor Reorder multiple-choice options to test whether answer position changes an LLM's selection despite unchanged question content.} \\
\bottomrule
\end{tabular}
\end{table}

\FloatBarrier
{The AutoLab tasks~\citep{xu2026autolabfrontiermodelssolve} test concrete implementation decisions. The agent starts from a working program and seeks better performance while satisfying correctness requirements. Table~\ref{tab:autolab_task_guide} distinguishes runtime optimization from the model-size objective in \tasknameinline{Smallest Game Player}.}

\begin{table}[!ht]
\centering
\caption{{\textbf{Optimization objectives of the AutoLab tasks.} These tasks span image processing, database operations, numerical computation, and compact model design. Each objective is assessed under the benchmark's correctness requirements.}}
\label{tab:autolab_task_guide}
\small
\setlength{\tabcolsep}{5pt}
\renewcommand{\arraystretch}{1.13}
\begin{tabular}{@{}>{\raggedright\arraybackslash}p{0.29\textwidth}>{\raggedright\arraybackslash}p{\dimexpr0.71\textwidth-10pt\relax}@{}}
\toprule
{\normalcolor \textbf{Task}} & {\normalcolor \textbf{What the agent optimizes}} \\
\midrule
{\normalcolor \taskname{Gaussian Blur}} & {\normalcolor Reduce the time required to smooth an image with a Gaussian filter.} \\
\addlinespace[4pt]
{\normalcolor \taskname{Hash Join}} & {\normalcolor Accelerate table joins that match rows by key, while preserving the correct joined output.} \\
\addlinespace[4pt]
{\normalcolor \taskname{Concurrent KV WAL}} & {\normalcolor Speed up concurrent operations in a key-value store with write-ahead logging, which records updates for recovery.} \\
\addlinespace[4pt]
{\normalcolor \taskname{Flash Attention}} & {\normalcolor Reduce the runtime of scaled dot-product attention through more efficient computation and memory access.} \\
\addlinespace[4pt]
{\normalcolor \taskname{FFT (Rust)}} & {\normalcolor Compute the discrete Fourier transform of a signal faster in Rust.} \\
\addlinespace[4pt]
{\normalcolor \taskname{VLIW Scheduler}} & {\normalcolor Group instructions into bundles for a very long instruction word processor to reduce execution cycles.} \\
\addlinespace[4pt]
{\normalcolor \taskname{Smallest Game Player}} & {\normalcolor Minimize a Connect-3 network's parameter count while maintaining at least $95\%$ accuracy against perfect-play decisions.} \\
\addlinespace[4pt]
{\normalcolor \taskname{Sha256 Throughput}} & {\normalcolor Increase SHA-256 hashing throughput while preserving correct outputs.} \\
\bottomrule
\end{tabular}
\end{table}

{MLE-Bench~\citep{mlebench} requires agents to develop prediction pipelines from the supplied competition data. Table~\ref{tab:mle_task_guide} explains the input, prediction target, and evaluation metric for each competition. The task names link to the benchmark's descriptions, and score directions match Table~\ref{tab:mlebench}.}

\begin{table}[!ht]
\centering
\caption{{\textbf{Prediction tasks and metrics in MLE-Bench.} The competitions cover medical imaging, materials properties, plant health, and authorship classification. RMSLE denotes root mean squared logarithmic error; ROC AUC measures the area under the receiver operating characteristic curve.}}
\label{tab:mle_task_guide}
\small
\setlength{\tabcolsep}{5pt}
\renewcommand{\arraystretch}{1.13}
\begin{tabular}{@{}>{\raggedright\arraybackslash}p{0.28\textwidth}>{\raggedright\arraybackslash}p{\dimexpr0.46\textwidth-10pt\relax}>{\raggedright\arraybackslash}p{\dimexpr0.26\textwidth-10pt\relax}@{}}
\toprule
{\normalcolor \textbf{Task}} & {\normalcolor \textbf{What the model predicts}} & {\normalcolor \textbf{Evaluation metric}} \\
\midrule
{\normalcolor \href{https://github.com/openai/mle-bench/blob/main/mlebench/competitions/aptos2019-blindness-detection/description.md}{\taskname{APTOS 2019 Blindness}}} & {\normalcolor Diabetic retinopathy severity on a five-level scale from retinal images.} & {\normalcolor Quadratic weighted kappa~$\uparrow$} \\
\addlinespace[4pt]
{\normalcolor \href{https://github.com/openai/mle-bench/blob/main/mlebench/competitions/nomad2018-predict-transparent-conductors/description.md}{\taskname{NOMAD 2018}}} & {\normalcolor Formation energy and bandgap energy of transparent-conductor materials from their composition and structure.} & {\normalcolor Mean RMSLE over the two targets $\downarrow$} \\
\addlinespace[4pt]
{\normalcolor \href{https://github.com/openai/mle-bench/blob/main/mlebench/competitions/plant-pathology-2020-fgvc7/description.md}{\taskname{Plant Pathology 2020}}} & {\normalcolor Whether an apple leaf is healthy or affected by rust, scab, or multiple diseases, using its photograph.} & {\normalcolor Mean ROC AUC across categories $\uparrow$} \\
\addlinespace[4pt]
{\normalcolor \href{https://github.com/openai/mle-bench/blob/main/mlebench/competitions/spooky-author-identification/description.md}{\taskname{Spooky Author ID}}} & {\normalcolor Which of three authors wrote a text excerpt: Edgar Allan Poe, Mary Shelley, or H.\,P. Lovecraft.} & {\normalcolor Multiclass log loss $\downarrow$} \\
\bottomrule
\end{tabular}
\end{table}